%% file: main.tex
\documentclass[10pt,twocolumn,letterpaper]{article}

\usepackage{cvpr}              

\input{preamble}

\definecolor{cvprblue}{rgb}{0.21,0.49,0.74}
\usepackage[pagebackref,breaklinks,colorlinks,citecolor=cvprblue]{hyperref}
\usepackage{nicematrix}
\usepackage{multirow}
\usepackage{booktabs}
\usepackage{mathtools}
\usepackage{soul}
\DeclarePairedDelimiter{\ceil}{\lceil}{\rceil}
\usepackage{cuted}
\usepackage{rotating}
\usepackage{algorithm}
\usepackage{algpseudocode}
\def\paperID{3210} 
\def\confName{CVPR}
\def\confYear{2024}
\def\modelname{ViewFusion}

\title{ViewFusion: Towards Multi-View Consistency via Interpolated Denoising}

\author{Xianghui Yang\textsuperscript{1,2}\thanks{Work done during internship at Amazon}, Yan Zuo\textsuperscript{1}, Sameera Ramasinghe\textsuperscript{1}, Loris Bazzani\textsuperscript{1}, Gil Avraham\textsuperscript{1}, Anton van den Hengel\textsuperscript{1,3}\\
\textsuperscript{1}Amazon, \textsuperscript{2}The University of Sydney, \textsuperscript{3}The University of Adelaide\\
}

\begin{document}
\maketitle


\input{sec/0_abstract}    
\input{sec/1_intro}
\input{sec/2_related}
\input{sec/3_methods}

\input{sec/5_experiments}
\input{sec/6_ablation}
\input{sec/7_conclusion}
{
    \small
    \bibliographystyle{ieeenat_fullname}
    \bibliography{main}
}

\input{sec/X_suppl}

\end{document}

%% file: preamble.tex
\usepackage[dvipsnames]{xcolor}
\usepackage{pifont}
\usepackage{adjustbox}
\newcommand{\red}[1]{{\color{red}#1}}


%% file: sec/0_abstract.tex
\begin{abstract}

Novel-view synthesis through diffusion models has demonstrated remarkable potential for generating diverse and high-quality images. Yet, the independent process of image generation in these prevailing methods leads to challenges in maintaining multiple view consistency. To address this, we introduce \modelname{}, a novel, training-free algorithm that can be seamlessly integrated into existing pre-trained diffusion models.
Our approach adopts an auto-regressive method that implicitly leverages previously generated views as context for next view generation, ensuring robust multi-view consistency during the novel-view generation process.
Through a diffusion process that fuses known-view information via interpolated denoising, our framework successfully extends single-view conditioned models to work in multiple-view conditional settings without any additional fine-tuning. Extensive experimental results demonstrate the effectiveness of~\modelname{} in generating consistent and detailed novel views.

\end{abstract}

%% file: sec/1_intro.tex
\section{Introduction}
\label{sec:intro}



Humans have a remarkable capacity for visualizing unseen perspectives from just a single image view -- an intuitive process that remains complex to model. Such an ability is known as Novel View Synthesis (NVS) and necessitates robust geometric priors to accurately infer three-dimensional details from flat imagery; lifting from a two-dimensional projection to a three-dimensional form involves assumptions and knowledge about the nature of the object and space. Recently, significant advancements in NVS have been brought forward by neural networks~\cite{yao2018mvsnet,tewari2020state,wang2021neus,mildenhall2020nerf,xie2022neural, meshrcnn, pixel2mesh, genmesh, atlasnet, sdfsrn}, where novel view generation for downstream reconstruction shows promising potential~\cite{zero123, watson2022novel}.

\begin{figure}
     \centering
     \begin{subfigure}[b]{0.49\linewidth}
         \centering
         \includegraphics[width=\linewidth]{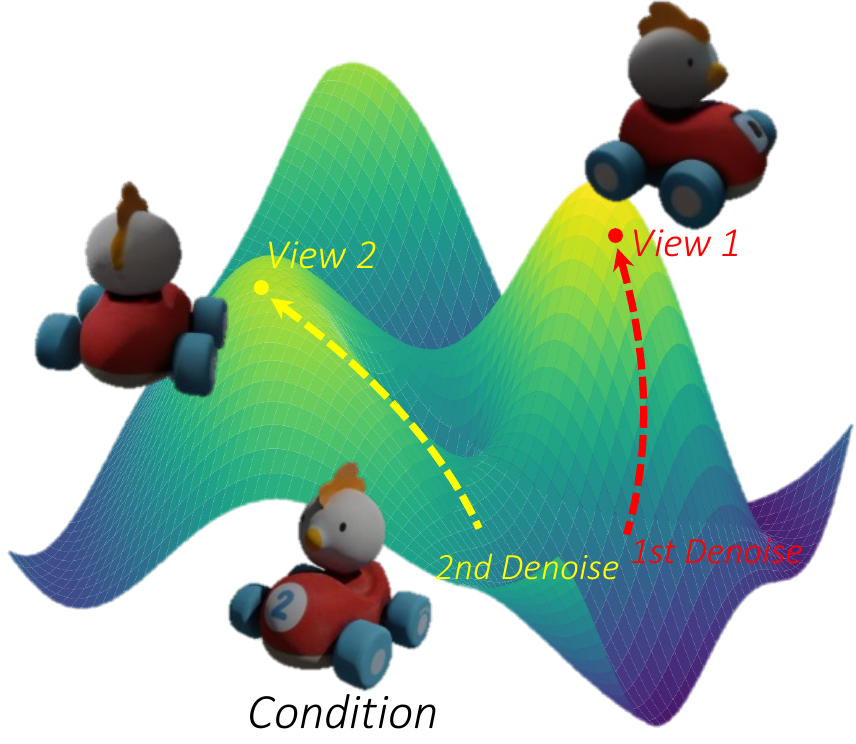}
         \caption{Inconsistent generation.}
         \label{fig:head_original}
     \end{subfigure}
     \hfill
     \begin{subfigure}[b]{0.49\linewidth}
         \centering
         \includegraphics[width=\linewidth]{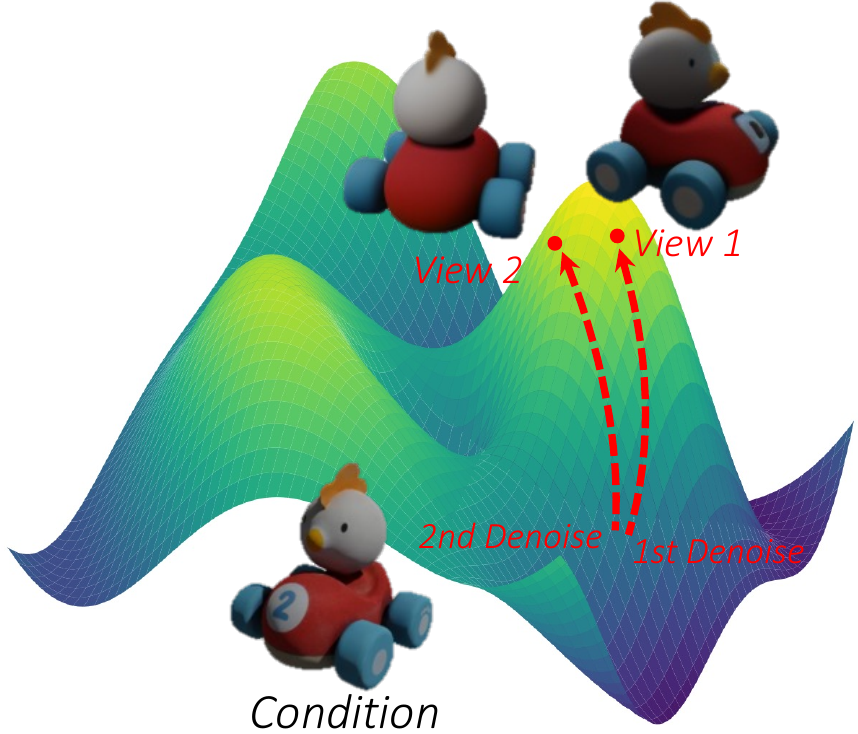}
         \caption{Consistent generation (ours).}
         \label{fig:head_auto}
     \end{subfigure}
     \vspace{-5pt}
    \caption{The cause of multi-view inconsistency in diffusion-based novel-view synthesis models. (a) Diffusion models incorporate randomness for diversity and better distribution modeling; this independent generation process produces realistic views under specific instances but may produce different plausible views for various instances, lacking alignment across adjacent views. (b) In contrast, \modelname{} incorporates an auto-regressive process to reduce uncertainty and achieve multi-view consistency, by ensuring a correlated denoising process that ends at the same high-density area, fostering consistency across views.}
  \label{fig:head}
  \vspace{-10pt}
\end{figure}



Specifically, diffusion models~\cite{rombach2022high,ho2020denoising} and their ability to generate high-quality 2D images have garnered significant attention in the 3D domain, where pre-trained, text-conditioned 2D diffusion models have been re-purposed for 3D applications via distillation~\cite{poole2022dreamfusion,wang2023score,wang2023prolificdreamer,lin2023magic3d,chen2023fantasia3d, tang2023make,melas2023realfusion,xu2022neurallift,raj2023dreambooth3d}. Follow-up approaches~\cite{watson2022novel, zero123} remove the requirement of text conditioning and instead take an image and target pose as conditions for NVS. However, distillation~\cite{wang2023score} is still required as the diffusion model cannot produce the multi-view consistent outputs that are appropriate for certain downstream tasks (\eg, optimizing Neural Radiance Fields (NeRFs)~\cite{mildenhall2020nerf}).

Under the single-view setting, maintaining multi-view consistency remains particularly challenging since there may exist several plausible outputs for a novel view that are aligned with the given input image. For diffusion-based approaches which generate novel views in an independent manner~\cite{watson2022novel, zero123}, this results in synthesized views containing artifacts of multi-view inconsistency (\cref{fig:head_original}). Previous work~\cite{one2345, syncdreamer, MVDream, ye2023consistent1to3, yang2023consistnet, long2023wonder3d} focuses on improving the robustness of the downstream reconstruction to address the inconsistency issue, including feature projection layers in the NeRF~\cite{one2345} or utilising three-dimensional priors to constrain NeRF optimization~\cite{ye2023consistent1to3, syncdreamer}, yet these techniques require training or fine-tuning to align additional modules to the original diffusion models.

In this work, we address the multi-view inconsistency that arises during the process of view synthesis. Rather than independently synthesizing views conditioned only on the initial reference image, we develop a novel approach where each subsequently generated view is also conditioned on the \emph{entire set} of previously generated views. Specifically, our method incorporates an auto-regressive process into the diffusion process to model the joint distribution of views, guiding our novel-view synthesis by maintaining the denoising direction towards the same high density area of already generated views (\cref{fig:head_auto}).

Our framework, named \modelname{}, relaxes the single-view conditioning requirement of typical diffusion models through an interpolated denoising process.~\modelname{} offers several additional advantages: 1) it can utilize all available views as guidance, thereby enhancing the quality of generated images by incorporating more information; 2) it does not require any additional fine-tuning, effortlessly converting pre-trained single-view conditioned diffusion models into multi-view conditioned diffusion models; 3) it provides greater flexibility in setting adaptive weights for condition images based on their relative view distance to the target view.

The contributions of this paper are the following:


\begin{itemize}
    \item We propose a \emph{training-free} algorithm which can be directly applied to pre-trained diffusion models to improve multi-view consistency of synthesized views and supports multiple conditional inputs.
    \item Our method utilizes a novel, auto-regressive approach which we call \textit{Interpolated Denoising}, that implicitly addresses key limitations of previous auto-regressive approaches for view synthesis.
    \item Extensive empirical analysis on ABO~\cite{abo} and GSO~\cite{downs2022google} show that our method is able to achieve better 3D consistency in image generation, leading to significant improvements in novel view synthesis and 3D reconstruction of shapes under single-view and multi-view image settings over other baseline methods.
\end{itemize}

%% file: sec/2_related.tex
\section{Related Work}
\subsection{3D-adapted Diffusion Models}
Diffusion models have excelled in image generation using conditional inputs \cite{reed2016generative, pix2pix2017, Park2017Semantic,zhang2023adding} and given this success in the 2D domain, recent works have tried to extend diffusion models to 3D content generation~\cite{nichol2022point,jun2023shap,muller2023diffrf,zhang20233dshape2vecset,liu2023meshdiffusion,wang2023rodin,gupta20233dgen,cheng2023sdfusion,karnewar2023holodiffusion,anciukevivcius2023renderdiffusion,zeng2022lion,erkocc2023hyperdiffusion,chen2023single,kim2023neuralfield,ntavelis2023autodecoding,gu2023learning,karnewar2023holofusion} -- although the scarcity of 3D data presents a significant challenge to directly train these diffusion models. Nonetheless, pioneer works such as DreamFusion~\cite{poole2022dreamfusion} and Score Jacobian Chaining~\cite{wang2023score} leverage pre-trained text-conditioned diffusion models to craft 3D models via distillation. Follow-up approaches~\cite{wang2023prolificdreamer,lin2023magic3d,chen2023fantasia3d,tang2023make} improve this distillation in terms of speed, resolution and shape quality. Approaches such as ~\cite{tang2023make,melas2023realfusion,xu2022neurallift,raj2023dreambooth3d} extend upon this to support image conditions through the use of captions with limited success due to the non-trivial nature of textual inversion~\cite{gal2022image}.




\subsection{Novel View Synthesis Diffusion Models}
Another line of research ~\cite{watson2022novel,gu2023nerfdiff,deng2023nerdi,zhou2023sparsefusion,tseng2023consistent,chan2023generative,yu2023long,tewari2023diffusion,yoo2023dreamsparse,szymanowicz2023viewset,tang2023mvdiffusion,xiang20233d,liu2023deceptive,lei2022generative} directly applies 2D diffusion models to generate multi-view images for shape reconstruction. To circumvent the weakness of text-conditioned diffusion models, novel-view synthesis diffusion models~\cite{watson2022novel, zero123} have also been explored, which take an image and target pose as conditions to generate novel views. However, for these approaches, recovering a 3D consistent shape is still a key challenge. To mitigate 3D inconsistency, \citet{one2345} suggests training a Neural Radiance Field (NeRF) with feature projection layers. Concurrently, other works~\cite{syncdreamer,ye2023consistent1to3,weng2023consistent123,yang2023consistnet,long2023wonder3d} add modules to original diffusion models for multi-view consistency, including epipolar attention~\cite{ye2023consistent1to3}, synchronized multi-view noise predictor~\cite{syncdreamer} and cross-view attention~\cite{weng2023consistent123,long2023wonder3d}; although these methods require fine-tuning an already pre-trained model. We adopt a different paradigm, instead of extending a single-view diffusion model with additional trainable models that incorporate multi-view conditions,
our training-free method enables pre-trained diffusion models to incorporate previously generated views via the denoising step and holistically extends these models into multi-view settings.

\subsection{Other Single-view Reconstruction Methods}
Before the prosperity of generative models used in 3D reconstruction, many works~\cite{tatarchenko2019single,fu2021single,kato2019learning,li2020self,fahim2021single,meshrcnn, pixel2mesh, genmesh, atlasnet,sdfsrn} reconstructed 3D shapes from single-view images using regression~\cite{li2020self,meshrcnn, pixel2mesh, genmesh, atlasnet} or retrieval~\cite{tatarchenko2019single}, both of which face difficulties in generalizing to real data or new categories. 
Methods based on Neural Radiance Fields (NeRFs)~\cite{mildenhall2020nerf} have found success in novel-view synthesis, but these approaches typically depend on densely captured images with accurately calibrated camera positions. Currently, several studies are investigating the adaptation of NeRF to single-view settings~\cite{pixelNeRF,lin2023visionnerf,jang2021codenerf,sargent2023vq3d}; although, reconstructing arbitrary objects from single-view images is still a challenging problem.

%% file: sec/3_methods.tex
\section{Method}
\begin{figure*}
\vspace{-8pt}
\centering
    \includegraphics[width=0.99\linewidth]{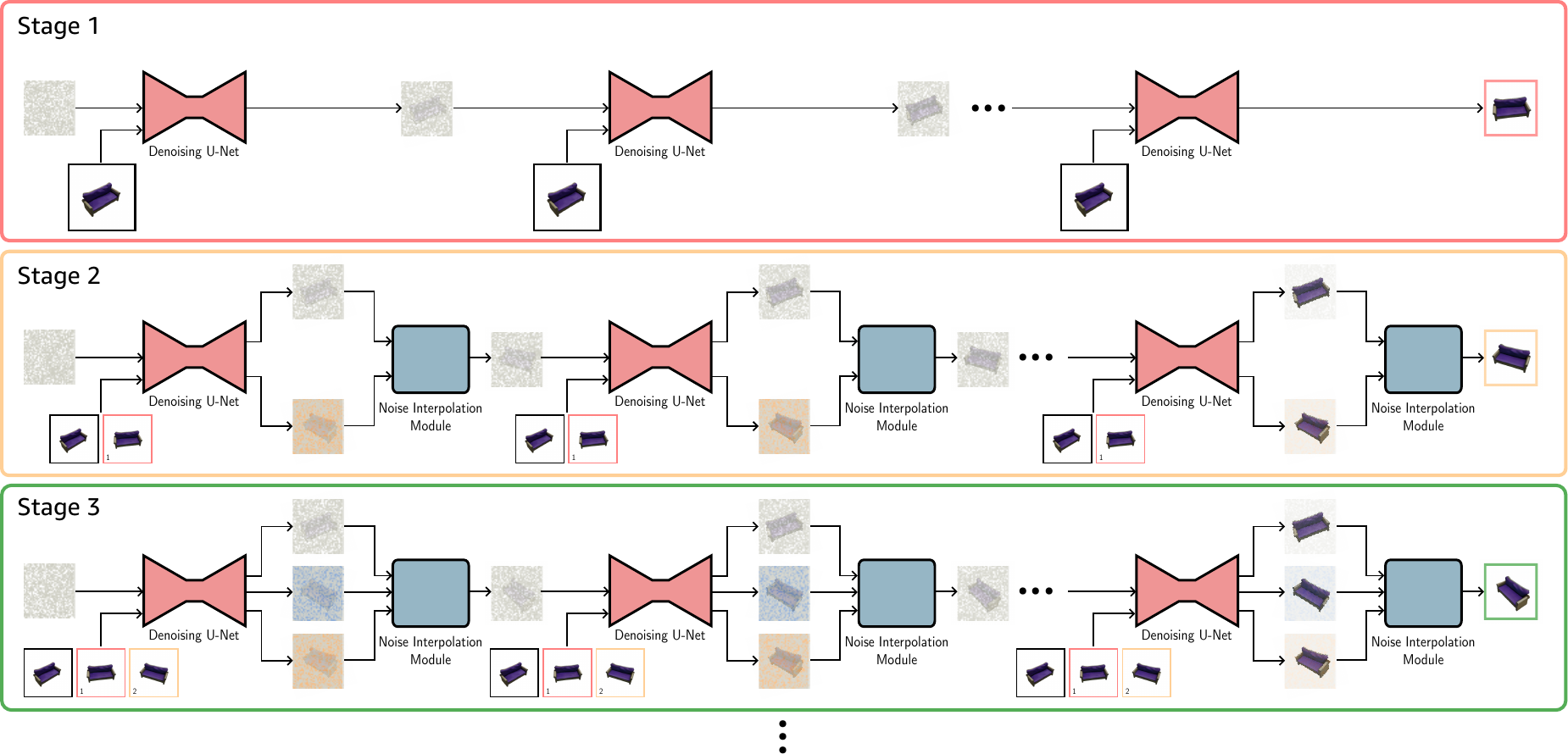}
\vspace{-8pt}
\caption{Illustration of the Auto-Regressive Generation Process. In our approach, we extend a pre-trained diffusion model from single-stage to multi-stage generation and we maintain a view set that contains all generated views. For each stage, we construct $N$ reverse diffusion processes and sharing a common starting noise. At each time step within this generation stage, the diffusion model predicts $N$ noises individually. These $N$ noises are then subjected to weighted interpolation through the \textit{Noise Interpolation Module}, concluding the denoising step with the a shared interpolated noise for subsequent denoising steps.}
  \label{fig:interpolation}
  \vspace{-5pt}
\end{figure*}

\subsection{Denoising Diffusion Probabilistic Models}
Denoising diffusion probabilistic models (DDPM)~\cite{sohl2015deep,ho2020denoising} are a class of generative models that model the real data distribution $q(x_0)$ with a tractable model distribution $p_\theta(x_0)$ by learning to iteratively denoise samples. It learns a probability model $ p_\theta(\mathbf{x}_0)=\int p_\theta (\mathbf{x}_{0:T})d\mathbf{x}_{1:T}$ to convert unstructured noise $\mathbf{x}_{T}$ to real samples $\mathbf{x}_0$ in the form of a Markov chain, with Gaussian transitions. The Gaussian transition is defined as:
\begin{equation}
q(\mathbf{x}_{T}|\mathbf{x}_0)=\prod_{t=1}^{T} q(\mathbf{x}_t|\mathbf{x}_{t-1})=
\prod_{t=1}^{T}\mathcal{N}(\mathbf{x}_t;\sqrt{1-\beta_t} \mathbf{x}_{t-1},\beta_t \mathbf{I}),
\label{eq:vanilla_forward}
\end{equation}
where $\beta_t, t\in\{1,...,T\}$ are the variance schedule parameter and timestep in the denoising process respectively. The reverse denoising process starts from a noise sampled from a Gaussian distribution $q(\mathbf{x}_T)=\mathcal{N}(\mathbf{0}, \mathbf{I})$ and is constructed as:
\begin{equation}
p_\theta(\mathbf{x}_{0}|\mathbf{x}_T)=\prod_{t=1}^{T} p_\theta(\mathbf{x}_{t-1}|\mathbf{x}_t)=\prod_{t=1}^{T}\mathcal{N}(\mathbf{x}_{t-1};\mathbf{\mu}_\theta(\mathbf{x}_t,t),\sigma^2_t \mathbf{I}),
\label{eq:vanilla_reverse}
\end{equation}
    
where the variance $\sigma^2_t$ is a time-dependent constant~\cite{ho2020denoising}, and $\mathbf{\mu}_\theta(\mathbf{x}_t,t)$ is the mean from the learned \textit{noise predictor} $\mathbf{\epsilon}_\theta$:
\begin{equation}
    \mathbf{\mu}_\theta(\mathbf{x}_t,t)=\frac{1}{\sqrt{\alpha}_t}\left(\mathbf{x}_t - \frac{\beta_t}{\sqrt{1-\bar{\alpha}_t}} \mathbf{\epsilon}_\theta (\mathbf{x}_t, t)\right).
    \label{eq:vanilla_mu}
\end{equation}
Here, $\alpha_t$ and $\bar{\alpha}_t$ are constants derived from $\beta_t$. The objective of noise predictor $\mathbf{\epsilon}_\theta$ is simplified to:
\begin{equation}
    \ell=\mathbb{E}_{t,\mathbf{x}_0,\mathbf{\epsilon}}\left[\|\mathbf{\epsilon} - \mathbf{\epsilon}_\theta (\sqrt{\bar{\alpha}_t} \mathbf{x}_0+\sqrt{1-\bar{\alpha}_t}\mathbf{\epsilon}, t)\|_2\right],
\end{equation}
where $\mathbf{\epsilon}$ is a random variable sampled from $\mathcal{N}(\mathbf{0},\mathbf{I})$~\cite{ho2020denoising}. 

\subsection{Pose-Conditional Diffusion Models}

Similar to other generative models~\cite{2014CoGAN, Sohn2015LearningSO}, diffusion models inherently possess the capability to model conditional distributions of the form $p_\theta(x_{t-1}|x_t, y)$ where $y$ is the condition. We employ a conditional denoising autoencoder, denoted as $\mathbf{\epsilon}_\theta(\mathbf{x}_t, t, y)$ which enables controlling the synthesis process through a variety of input modalities, including textual descriptions~\cite{reed2016generative}, semantic maps~\cite{pix2pix2017, Park2017Semantic}, or other image-to-image translation tasks~\cite{pix2pix2017}. In the following, we present a range of approaches to novel-view synthesis, exploring how various works, including our own, approach the concept of a single reverse diffusion step. Through this comparison, we clarify and establish the underlying relationships between these different methodologies. The notation will follow that bottom subscript $(\cdot)_{t}$ indicates the diffusion step and upper subscript $(\cdot)^{i}$ relates to the view index. Subsequently, the $i$-th condition image and its relative pose to the target view are defined as $\mathbf{y}^i$ and $\pi^i$, respectively, and the noisy image to be denoised at timestep $t$ is defined as $\mathbf{x}_t$. 

\paragraph{Direct condition} was applied by Zero 1-to-3~\cite{zero123} to the reverse process when given a single input image and target pose $\mathbf{y}^{1}, \pi^{1}$:
\begin{equation}
    p(\mathbf{x}_{t-1}|\mathbf{x}_{t}, \mathbf{y}^{1}, \pi^{1}).
\end{equation}

\paragraph{Stochastic conditioning} was formulated by~\cite{watson2022novel} which can leverage multiple views sampled from a collection of views $p_{\mathbf{y},\pi}(\mathcal{Y}, \mathcal{\pi})$:
\begin{equation}
    p(\mathbf{x}_{t-1}|\mathbf{x}_{t}, \mathbf{y}^{i}, \pi^{i}),  \{\mathbf{y}^{i}, \pi^{i}\} \sim p_{\mathbf{y},\pi}(\mathcal{Y}, \mathcal{\pi}),
\end{equation}
where the sampling of image and pose happens at each diffusion step $t$. \newline

\paragraph{Joint output distribution} was shown in SyncDreamer\cite{syncdreamer} which learns a joint distribution of many views given an image condition $y^{1}$:
\begin{equation}
    p(\mathbf{x}_{t-1}^{1:N}|\mathbf{x}_{t}^{1:N}, \mathbf{y}^{1}, e^{1}),
\end{equation}
where $N$ is the number of generated novel views and $e$ is the elevation condition (partial pose information). We note that in this formulation the target poses are not fully specified as part of the condition allowing for diverse pose generation of outputs. \newline

\paragraph{Auto-regressive distribution} is an auto-regressive distribution setting which can generate an arbitrary number of views given a single or multiple condition images and poses contained in the set of $\mathbf{y}^{1:N-1}, \pi^{1:N-1}$:
\begin{equation}
    p(\mathbf{x}_{t-1}^{N}|\mathbf{x}_{t}^{N}, \mathbf{y}^{1:N-1}, \pi^{1:N-1}).
\end{equation}
Our approach falls in the auto-regressive category and for the remainder of this section we detail the implementation to achieve this sampling strategy.


\subsection{Interpolated Denoising}

The standard DDPM model has been adapted for novel-view image synthesis by using an image and target pose (\ie, rotation and translation offsets) as conditional inputs~\cite{watson2022novel}. Following training on a large-scale dataset, this approach has demonstrated the capability for zero-shot reconstruction~\cite{zero123}. To address the challenge of maintaining multi-view consistency, we employ an auto-regressive approach for generating sequential frames (See ~\cref{fig:interpolation}). Instead of independently producing each frame from just the input images -- a process prone to significant variations between adjacent images -- we integrate an auto-regressive algorithm into the diffusion process. This integration enables us to model a conditional joint distribution, ensuring smoother and more consistent transitions between frames.

To guide the synthesis of novel views using images under different views, we design an interpolated denoising process. For the purpose of this derivation, we assume access to an image set containing $N-1$ images denoted as $\{\mathbf{y}^{1},...,\mathbf{y}^{N-1}\}$. We want to model the distribution of the $N$-th view image conditioned on these $N-1$ views $q(\mathbf{x}_{1:T}^{N}|{\mathbf{y}^{1:N-1}})$,
where the relative pose offsets $\pi^{i}, i\in\{1,N-1\}$ between the condition images $\{\mathbf{y}^{1},...,\mathbf{y}^{N-1}\}$ and target image $\mathbf{x}^{N}_0$ are omitted for simplicity. The forward process of the multi-view conditioned diffusion model is a direct extension of the vanilla DDPM in Eq.~\ref{eq:vanilla_forward}, where noises are added to every view independently by
\begin{equation}
\label{eq:mv_forward}
\begin{split}
    q(\mathbf{x}^{N}_{1:T}|\mathbf{y}^{1:N}) & = \prod_{t=1}^T q(\mathbf{x}^{N}_{t}|\mathbf{x}_{t-1}^{N},\mathbf{y}^{1:N})
\end{split}
\end{equation}
where $q(\mathbf{x}^{N}_t|\mathbf{x}^{N}_{t-1},\mathbf{y}^{1:N})=\mathcal{N}(\mathbf{x}^{N}_t;\sqrt{1-\beta_t} \mathbf{x}^{N}_{t-1},\beta_t \mathbf{I})$. The initial is defined as $\mathbf{x}_{0}^{N} := \mathbf{y}^{N}$.
Similarly, following Eq.~\ref{eq:vanilla_reverse}, the $\log$ reverse process is constructed as
\begin{equation}
\label{eq:mv_backward}
\begin{aligned}
\log p_\theta(\mathbf{x}_{0}^N|\mathbf{x}_T^N,&\mathbf{y}^{1:N-1})=\sum_{t=1}^{T} \log p_\theta(\mathbf{x}_{t-1}^N|\mathbf{x}_t^N,\mathbf{y}^{1:N-1})\\
\underset{(1)}{\approx}&\sum_{t=1}^{T} \log \prod_{n=1}^{N-1}p_\theta(\mathbf{x}_{t-1}^N|\mathbf{x}_t^N,\mathbf{y}^n)\\
=&\sum_{t=1}^{T}\sum_{n=1}^{N-1} \log \mathcal{N}(\mathbf{x}_{t-1}^N;\mathbf{\mu}_\theta^n(\mathbf{x}_t^N,\mathbf{y}^n,t),\sigma^2_t \mathbf{I})\\
=&\sum_{t=1}^{T}\mathcal{N}\left(\mathbf{x}_{t-1}^N;\mathbf{\bar{\mu}}_\theta(\mathbf{x}_t^N,\mathbf{y}^{1:N-1},t),\bar{\sigma_t}^2 \mathbf{I}\right).
\end{aligned}
\end{equation}
Where $\bar{\mu_{\theta}}, \bar{\sigma_t}^2$ are taken as the mean and variance of the summation of $N-1$ \textit{log}-normal distributions. A note on subscript $(1)$ in Eq.\ref{eq:mv_backward}; to avoid cluttering the derivation, we assume $N-1$ independent inferences of the same random variable $\mathbf{x}_{t-1}^{N}$ using \textit{a different} $\mathbf{y}^{n}$ that results in $N-1$ independent normal distributions, which would require an additional subscript that we omitted for clarity. 
\subsection{Single and Multi-view Denoising}
\textit{In practice}, however, we may not have all $N-1$ views but a single view or a handful of views. For the reminder of this section, we treat an estimated view as $\mathbf{x}_{0}^{n}$, to be the $n$-th view $\mathbf{y}^{n}$ after a full reverse diffusion process. We use $\bar{\mathbf{\mu}}_\theta(\mathbf{x}_t,\mathbf{y}^{1:N-1},t)$ as the weighted average of $\mathbf{\mu}_\theta^n(\mathbf{x}_t,\mathbf{y}^n,t)$. For computing $\bar{\mu}_{\theta}$ using both given views and estimated views we adopt an approach where different views contribute differently to the target view, and we assign the weight $\omega_n$ for the $n$-th view in practice while satisfying the constraint $\sum_{n=1}^{N-1}{w_n}=1$. The \textit{Noise Interpolation Module} in~\cref{fig:interpolation} is modeled as:
\begin{equation}
\begin{aligned}
    \bar{\mathbf{\mu}}_\theta(\mathbf{x}_t,\mathbf{y}^{1:N-1},&t)=\sum_{n=1}^{N-1}\omega_n\mathbf{\mu}_\theta^n(\mathbf{x}_t, \mathbf{y}^{n}, t)\\
    =&\sum_{n=1}^{N-1}\omega_n\frac{1}{\sqrt{\alpha}_t}\left(\mathbf{x}_t - \frac{\beta_t}{\sqrt{1-\bar{\alpha}_t}} \mathbf{\epsilon}_\theta (\mathbf{x}_t, \mathbf{y}^{n}, t)\right)\\
    =&\frac{1}{\sqrt{\alpha}_t}\left(\mathbf{x}_t - \frac{\beta_t}{\sqrt{1-\bar{\alpha}_t}} \sum_{n=1}^{N}\omega_n\mathbf{\epsilon}_\theta (\mathbf{x}_t, \mathbf{y}^{n}, t)\right).\\
\end{aligned}
\end{equation}

In our approach, as the full view set is not given to us, we approximate this process by an auto-regressive way and grow the condition set during the generation. We define the weight parameter $\omega_n$ based on the angle offset, ~\ie, azimuth ($\Delta_{a}^{n}$), elevation ($\Delta_{e}^{n}$), and distance ($\Delta_{d}^{n}$), between the target view and the $n-th$ condition view. The core idea is to assign higher importance to near-view images during the denoising process while ensuring that the weight for the initial condition image does not diminish too rapidly, even when the target view is positioned at a nearly opposite angle. 
We use an exponential decay weight function for the initial condition image, defined as $\omega_{n}=e^{-\frac{\Delta^{n}}{\tau_c}}$.
Here, $\tau_c$ is the temperature parameter that regulates the decay speed, and $\Delta^{n}$ is the sum of the absolute relative azimuth ($\Delta_{a}^{n}$), elevation ($\Delta_{e}^{n}$), and distance ($\Delta_{d}^{n}$) between the target and condition poses. We calculate $\Delta^{n}$ as $\Delta^{n}=|\Delta_{a}^{n}|/\pi+|\Delta_{e}^{n}|/\pi+|\Delta_{d}^{n}|$. 

For the weights of the remaining images denoted as $\{x_0^2, ..., x_0^N\}$, all generated from the initial condition image $y^1 := x_0^1$, we use a softmax function to define the weights $\omega_n$:
\begin{equation}
\label{eq:generation_weights}
\begin{split}
    \omega_n=\text{Softmax}(\frac{e^{-\frac{\Delta^{n}}{\tau_g}}}{\sum_{n=2}^N e^{-\frac{\Delta^n}{\tau_g}}}), n=2,...,N
\end{split}
\end{equation}
Similarly, $\Delta_{n}$ represents the relative pose offset between target view and the $n$-th generated view,  and $\tau_g$ represents the temperature parameter for generated views. 
As an example, in the single-view case, the weights are expressed as follows,
\begin{equation}
\label{eq:weights}
\omega_n=\left\{
\begin{aligned}
    &exp(-\frac{\Delta^{n}}{\tau_c}), & n=1\\
    &(1-\omega_1)\text{Softmax}(\frac{e^{-\frac{\Delta^{n}}{\tau_g}}}{\sum_{n=2}^N e^{-\frac{\Delta^n}{\tau_g}}}), & n\neq 1
\end{aligned}
\right.
\end{equation}
we apply the term $1-\omega_1$ on the generated image weights to ensure the requirement of $\sum_{n=1}^{N-1}{w_n}=1$ will be met. In practice, Eq.\ref{eq:weights} is generalised to allow the condition set can be larger than $1$, \ie, multi-view generation (see supplementary).

\subsection{Step-by-step Generation}
\begin{figure}
     \centering
     \begin{subfigure}[b]{0.45\linewidth}
         \centering
         \includegraphics[width=\linewidth]{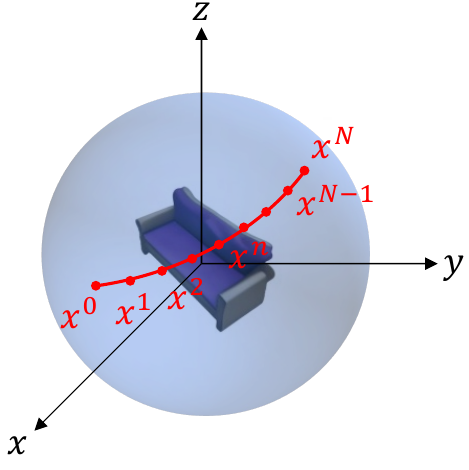}
         \caption{Single image generation.}
         \label{fig:single_trajectory}
     \end{subfigure}
     \hfill
     \begin{subfigure}[b]{0.45\linewidth}
         \centering
         \includegraphics[width=\linewidth]{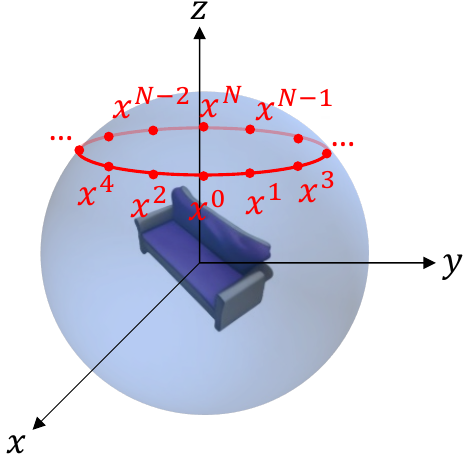}
         \caption{Spin video generation.}
         \label{fig:spin}
     \end{subfigure}
    \vspace{-8pt}
    \caption{Illustration of Step-by-step Generation. (a) we uniformly sample views along this trajectory in sequence to generate a novel-view image; (b) we sample views from nearest to furthest views according to to view distance to generate a $360^{\circ}$ spin video.}
    \vspace{-8pt}
\end{figure}

\textbf{Single image generation.} When applying the auto-regressive approach to image generation, we have devised a generation trajectory, as illustrated in~\cref{fig:single_trajectory}. We uniformly sample views along this trajectory in sequence. Each previously generated view image on this trajectory is incorporated into the condition set, providing guidance for the subsequent denoising process via our interpolated denoising method. To determine the number of steps, denoted as $S$, needed for this trajectory, we use the following formula:
\begin{equation}
S = max\left(\ceil{\frac{\Delta_{a}^N}{\delta}}, \ceil{\frac{\Delta_e^N}{\delta}}\right).
\end{equation}
Here, we set the maximum offset per step $\delta$ to determine the step count $S$, also based on the target view offsets $\Delta_{a}^N$ and $\Delta_e^N$. We then proceed to sample the $n$-th view using the following equation:
\begin{equation}
(\Delta_a^n, \Delta_e^n, \Delta_d^n)=(\frac{\Delta_a^N}{S}*n, \frac{\Delta_e^N}{S}*n, \frac{\Delta_d^N}{S}*n)
\end{equation}

\noindent \textbf{Spin videos generation.} In contrast to generating a single target image, the process of spin video generation begins from an initial image and concludes at the same position. To achieve this, we need to modify the generation order to leverage the broad range of rotation images, rather than simply following the rotation degree range of $[0^\circ,360^\circ]$ in sequence. This is because, at $\Delta_a=\pi$, the view is opposite to the conditioning view, marking the end of the generation process. To establish the generation order for spin video generation, we introduce the minimum azimuth offset, denoted as $\delta$, and employ a skip trajectory with the following order: $\{\delta, -\delta, 2\delta, -2\delta..., N\delta\}$, shown in~\cref{fig:spin}. For simplicity, we only consider rotation along the azimuth dimension in this context.

%% file: sec/5_experiments.tex
\section{Experiments}
\paragraph{Datasets.} We evaluate our method and compare to baselines on the ABO~\cite{abo} and GSO~\cite{downs2022google} datasets. These datasets are out-of-the-distribution as all baselines are trained on the Objaverse~\cite{deitke2023objaverse}. We also provide qualitative results on real images to showcase performance of our method on in-the-wild images in the supplementary. For additional results, please refer to the videos contained in the supplementary.


\paragraph{Metrics.} We assess our novel-view synthesis on three main criteria:
\begin{enumerate}
    \item \emph{Image Quality}: LPIPS~\cite{zhang2018unreasonable}, PSNR, and SSIM~\cite{ssim} metrics to help gauge the similarity between synthesized and ground-truth views.
    \item \emph{Multi-View Consistency}: Using SIFT~\cite{SIFT}, LPIPS~\cite{zhang2018unreasonable} and CLIP~\cite{radford2021learning}, we measure the uniformity of images across various perspectives.
    \item \emph{3D Reconstruction}: Chamfer distances and F-score between ground-truth and reconstructed shapes determine geometrical consistency.
\end{enumerate}


\begin{table}[t]
  \centering
  \resizebox{\linewidth}{!}{
  \begin{NiceTabular}{@{}lcccc|ccc@{}}[code-before =
  \cellcolor{gray!10}{4-3,4-4,4-5} 
  \cellcolor{gray!10}{7-3,7-4,7-5} 
  ]
    \toprule
    \multirow{2}{3em}{Dataset} & \multirow{2}{3em}{Method} & \multicolumn{3}{c}{Free Renderings} & \multicolumn{3}{c}{SyncDreamer Renderings} \\
    \cline{3-8}
     &  & SSIM$\uparrow$ & PSNR$\uparrow$ & LPIPS$\downarrow$  & SSIM$\uparrow$ & PSNR$\uparrow$ & LPIPS$\downarrow$ \\
    \midrule
    \multirow{3}{3em}{ABO} & Zero123~\cite{zero123} & 0.8796 & 21.33 & 0.0961 & 0.7822 & 18.27 & 0.1999\\
     & SyncDre.~\cite{syncdreamer} & \textcolor{gray!80}{0.7712} & \textcolor{gray!80}{13.43} & \textcolor{gray!80}{0.2182} & \textbf{0.8031} & \textbf{19.07} & \textbf{0.1816}\\
     & Ours & \textbf{0.8848} & \textbf{21.43} & \textbf{0.0923} & 0.7983 & 18.75 & 0.1985\\
    \bottomrule
    \multirow{3}{3em}{GSO} & Zero123~\cite{zero123} & 0.8710 & 20.33 & 0.1029 & 0.7925 & 18.06 & 0.1714\\
     & SyncDre.~\cite{syncdreamer} & \textcolor{gray!80}{0.8023} & \textcolor{gray!80}{14.42} & \textcolor{gray!80}{0.1833} & 0.8024 & 18.20 & \textbf{0.1647}\\
    & Ours & \textbf{0.8820} & \textbf{20.73} & \textbf{0.0958} & \textbf{0.8076} & \textbf{18.40} & 0.1703\\
    \bottomrule
  \end{NiceTabular}
  }
  \vspace{-8pt}
    
  \caption{Quantitative results on ABO and GSO datasets with \textbf{arbitrary} (left) and \textbf{discrete} (right) rotation and translation. Free renderings are a set of arbitrary rotation and translation as target generation view, while SyncDreamer renderings are a fixed set of 16 views with discrete azimuth, fixed elevation and distance, \ie, azimuth$\in\{0^\circ, 22.5^\circ, 45^\circ, ..., 315^\circ, 337.5^\circ\}$, elevation=$30^\circ$. Note that SyncDreamer~\cite{syncdreamer} \textit{cannot} generate images under arbitrary views apart from the predefined 16 camera positions.}
  
  \label{tab:novel_view}
\end{table}

\begin{figure*}[t]
\vspace{-8pt}
\centering
    \includegraphics[width=0.94\linewidth]{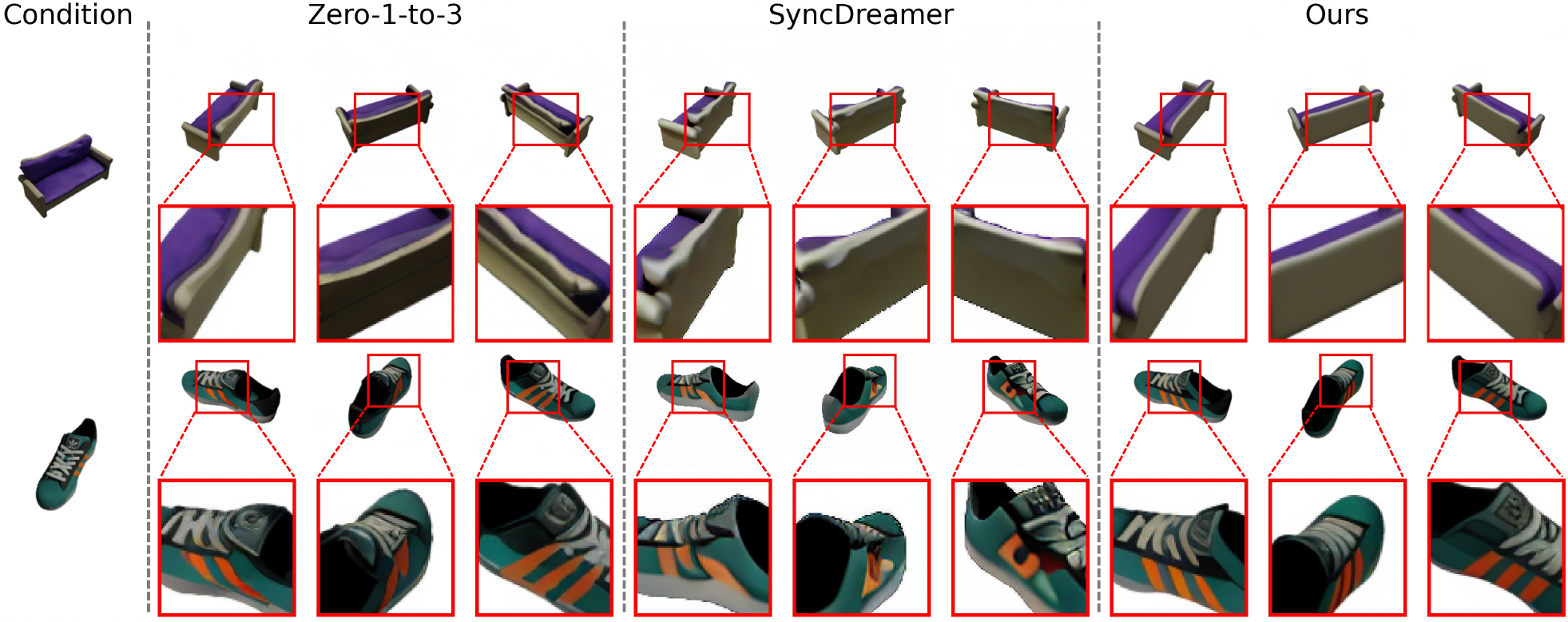}
\vspace{-8pt}
  \caption{Qualitative results for $360^\circ$ Spin Video Generation. Note the additional consistency in generated views our approach offers over the competing baselines shown in the bounding boxes.}
  \label{fig:consistency}
  \vspace{-8pt}
\end{figure*}

\subsection{Novel-view synthesis}

In~\cref{tab:novel_view}, we show quantitative results for novel-view synthesis under arbitrary and fixed-view settings. The fixed-view setting uses the rendering set of~\cite{syncdreamer} and ensures a fair comparison to~\cite{syncdreamer} which is limited to this fixed-view generation setting. As shown in~\cref{tab:novel_view}, our method can produce comparable results to~\cite{syncdreamer}, without any fine-tuning on the rendering set of~\cite{syncdreamer}. Under the arbitrary-view setting, we sample the nearest view from the rendering set of~\cite{syncdreamer} to the designated target view for generation;~\cite{syncdreamer} under-performs both~\cite{zero123} and our approach. Overall, regardless of the evaluation setting, our method consistently outperforms~\cite{zero123} and even outperforms~\cite{syncdreamer} on the GSO dataset under the favourable fixed-view setting for~\cite{syncdreamer}.

The task of novel view synthesis serves as a precursor for enabling more significant downstream applications, such as 3D reconstruction, by generating requisite input views. Thus, in addition to image quality, thorough evaluation needs to encompass the multi-view consistency of the synthesized perspectives. This additional criterion ensures that the generated imagery not only appears visually compelling but also aligns geometrically across different viewpoints. In the following sections, we evaluate against baselines on 3D consistency and show that the significant improvements our approach offers.

\paragraph{Multi-view Consistency.}

\begin{figure}[t]
    \includegraphics[trim={2cm 1cm 0 0}, clip, width=0.99\linewidth]{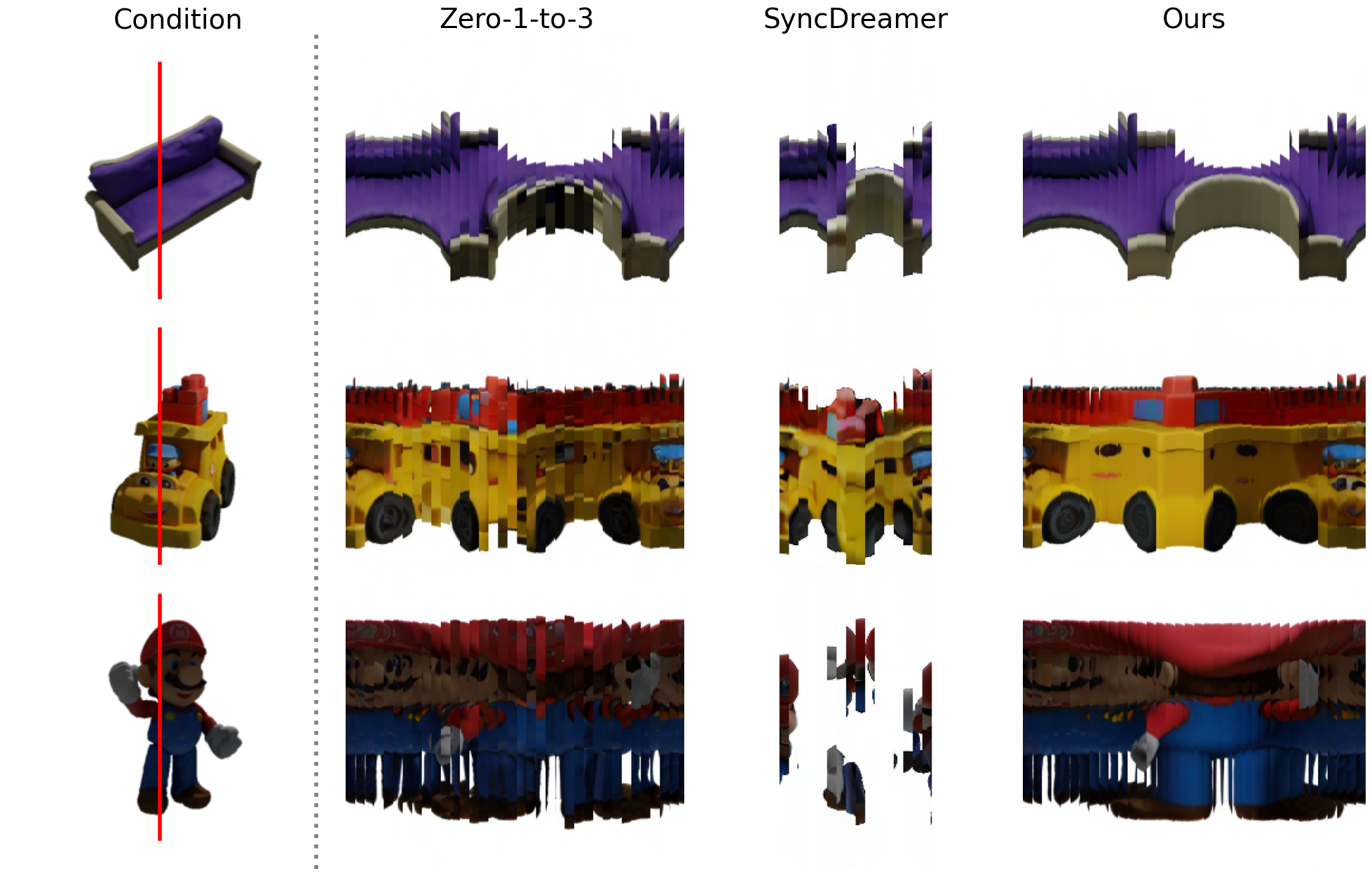}
\vspace{-8pt}
  \caption{Qualitative comparison for Motion Smoothness. We visualize the output videos using space-time Y-t slices through frames of the generated spin video (along the scanline shown in the condition).}
  \vspace{-8pt}
  \label{fig:motion}
\end{figure}

For measuring multi-view consistency of synthesized views, quantitative results are shown in~\cref{tab:consistency}, while qualitative comparisons are shown in~\cref{fig:consistency,fig:motion}. Here, we can see when compared to the baselines of~\cite{zero123} and~\cite{syncdreamer}, our approach excels in generating images that are both semantically consistent with the input image and maintains multi-view consistency in terms of colors and geometry under arbitrary-view settings.

\subsection{Multi-view conditional setting for NVS.} Additionally, our approach allows for the extension of single-view conditioned models into multi-view conditioned models which can take multiple conditional images as input. The results presented in~\cref{tab:multi_view} showcase the advantages our method offers, as novel-view synthesis quality improves with an increasing number of conditional input views. This is a significant improvement over the Zero-1-to-3 baseline and demonstrates the efficacy of our proposed method in the multi-view setting.

\begin{table}
  \centering
  \resizebox{0.85\linewidth}{!}{
  \begin{tabular}{@{}lccccc@{}}
    \toprule
    Dataset & Method & View & SIFT$\uparrow$ & LPIPS$\downarrow$ & CLIP$\uparrow$ \\
    \midrule
    \multirow{5}{*}{ABO} & Zero123~\cite{zero123} & \multirow{3}{*}{16}  & 13.38 & 0.1782 & 0.9604 \\
     & SyncDreamer~\cite{syncdreamer} &    & 12.36 & 0.1895 & 0.9584 \\
     & Ours &   & \textbf{13.51} & \textbf{0.1602} & \textbf{0.9664} \\
     \cline{2-6}
    & Zero123~\cite{zero123} & \multirow{2}{*}{36} & 17.03 & 0.1231 & 0.9725\\
     & Ours &    & \textbf{18.01} & \textbf{0.0966} & \textbf{0.9812}\\
    \bottomrule
    \multirow{5}{*}{GSO} & Zero123~\cite{zero123} & \multirow{3}{*}{16}   & 12.58 & 0.1411 & 0.9482\\
    & SyncDreamer~\cite{syncdreamer} &   & 13.24 & 0.1315 & 0.9532 \\
    & Ours &   & \textbf{13.83} & \textbf{0.1187} & \textbf{0.9601}  \\
    \cline{2-6}
    & Zero123~\cite{zero123} & \multirow{2}{*}{36} & 15.20 & 0.1056 & 0.9592 \\
    & Ours &   & \textbf{17.95} & \textbf{0.0676} & \textbf{0.9773} \\
    \bottomrule
  \end{tabular}
  }
  \vspace{-8pt}
  \caption{Quantitative results for multi-view consistency. We report the SIFT matching point number, LPIPS and CLIP similarity between adjacent frames to evaluate the multi-view consistency. Note that SyncDreamer can only generate 16 view images for a spin video due to the constraints imposed by its training.}
  \label{tab:consistency}
  \vspace{-8pt}
\end{table}

\begin{table}
  \centering
  \resizebox{0.85\linewidth}{!}{
  \begin{tabular}{@{}lcccc@{}}
    \toprule
    Dataset & Method & SSIM$\uparrow$ & PSNR$\uparrow$ & LPIPS$\downarrow$ \\
    \midrule
    \multirow{4}{*}{ABO} & Zero123~\cite{zero123} (1 view) & 0.8820 & 21.51 & 0.0945\\
    & Ours (1 view) & 0.8870 & 21.61 & 0.0904 \\
     & Ours (2 views) & 0.8913 & 21.92 & 0.0887 \\
     & Ours (3 views) & \textbf{0.8995} & \textbf{22.74} & \textbf{0.0815} \\
     \midrule
    \multirow{4}{*}{GSO} & Zero123~\cite{zero123} (1 view) &0.8721 & 20.42 & 0.1017\\
    & Ours (1 view) & 0.8830 & 20.87 & 0.0948 \\
     & Ours (2 views) & 0.8901 & 21.25 & 0.0893\\
     & Ours (3 views) & \textbf{0.8979} & \textbf{21.95} & \textbf{0.0792}\\
    \bottomrule
  \end{tabular}
  }
  \vspace{-8pt}
  \caption{
Quantitative results for multi-conditioned generation. Our approach outperforms the Zero-1-to-3 baseline~\cite{zero123} by extending its single-view reconstruction framework to a multi-view reconstruction framework.  Note that 3 condition views are removed from test set, and thus the results are slightly different to~\cref{tab:novel_view}.
  }
  \label{tab:multi_view}
  \vspace{-8pt}
\end{table}

\subsection{3D Reconstruction}
\begin{figure*}[t]
    \includegraphics[trim={0 1cm 0 0}, clip, width=0.99\linewidth]{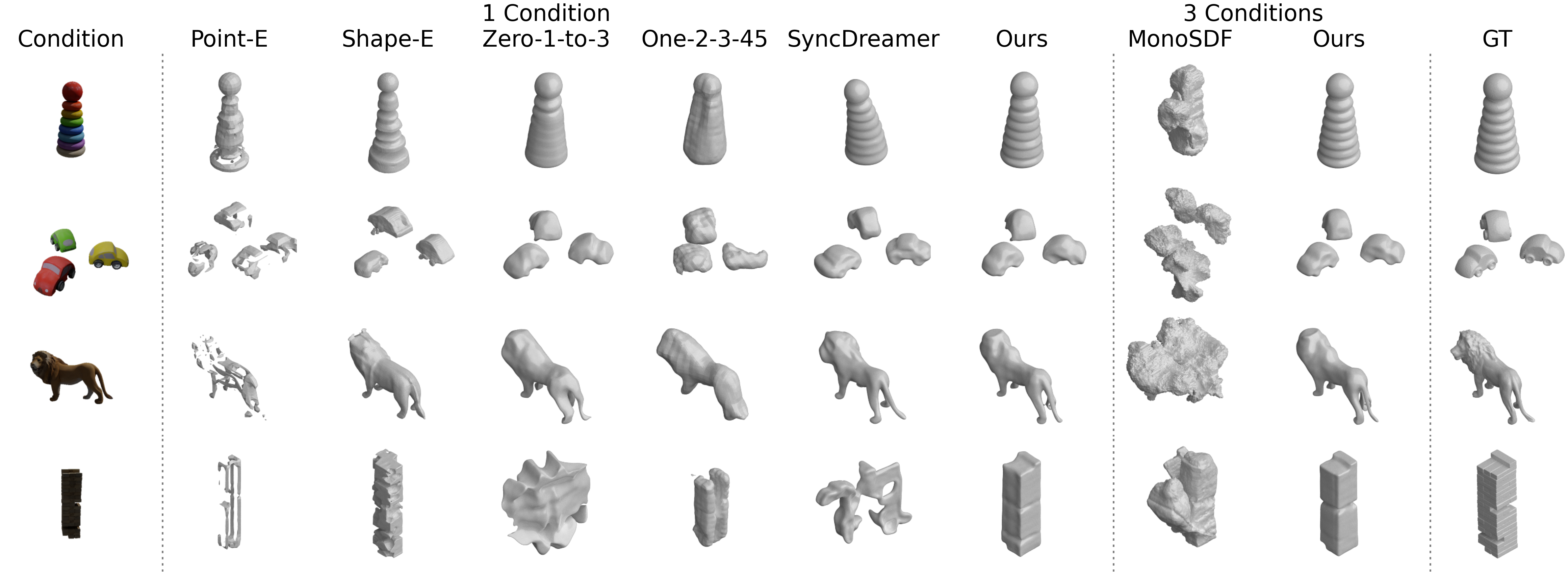}
  \vspace{-8pt}
  \caption{Qualitative results for single-view (1 condition image) and multi-view (3 condition images) 3D Reconstruction. Our method offers the most consistent results across a variety of different reconstructed shapes.}
  \label{fig:shape}
  \vspace{-8pt}
\end{figure*}

For shape reconstruction, we present our results both quantitatively (\cref{tab:shape}) and qualitatively (\cref{fig:shape}). We compare against baselines which synthesize novel views as well as direct image-to-shape approaches~\cite{nichol2022point,jun2023shape}. For the former approach, instead of using distillation~\cite{wang2023score}, we train NeuS~\cite{wang2021neus} on synthesized images to recover a shape.

\paragraph{Single-view Reconstruction.} We first compare our proposed method with several baselines under the single-view reconstruction setting, including Point-E~\cite{nichol2022point} and Shap-E~\cite{jun2023shape}, Zero-1-to-3~\cite{zero123}, One-2-3-45~\cite{one2345} and SyncDreamer~\cite{syncdreamer}. Qualitative results are shown in~\cref{fig:shape}; Point-E and Shap-E tend to generate incomplete shapes due to the single-view setting. Multi-view alignment methods such as Syncdreamer~\cite{syncdreamer} and One-2-3-45~\cite{one2345} capture the general geometry but tend to lose fine details. In comparison, our proposed method achieves the highest reconstruction quality amongst all approaches, where we can generate smooth surfaces and capture detailed geometry with precision. 

\begin{table}
  \centering
  \resizebox{\linewidth}{!}{
  \begin{NiceTabular}{@{}c|cccc|cc@{}}
    \toprule
    \multirow{2}{4em}{\centering Cond. views} & \multirow{2}{4em}{Method} & \multirow{2}{4em}{\centering Gen. views} & \multicolumn{2}{c|}{ABO} & \multicolumn{2}{c}{GSO} \\
    \cline{4-7}
    & & & CD$\downarrow$ & F-score$\uparrow$ & CD$\downarrow$ & F-score$\uparrow$ \\
    \midrule
    \multirow{8}{4em}{\centering 1}
    & Point-E~\cite{nichol2022point}  & N/A & 0.0428 & 0.7144 & 0.0672 & 0.6340 \\
    & Shap-E~\cite{jun2023shape}  & N/A & 0.0466 & 0.7364 & 0.0384 & 0.7313 \\
    & One-2-3-45~\cite{one2345}  & 32 & 0.0419 & 0.6665 & 0.0408 & 0.6490 \\
    & SyncDreamer~\cite{syncdreamer}  & 16 & 0.0160 & 0.8187 & 0.0229 & 0.7767 \\
    & Zero123~\cite{zero123}  & 16 & 0.0147 & 0.8226 & 0.0206 & 0.8045  \\
    & Zero123~\cite{zero123}  & 36 & 0.0139 & 0.8247 & 0.0207 & 0.8078 \\
    & Ours  & 16 & 0.0133 & 0.8423 & 0.0177 & 0.8274 \\
    & Ours  & 36 & \textbf{0.0126} & \textbf{0.8472} & \textbf{0.0164} & \textbf{0.8436} \\
    \cline{1-7}
    \multirow{2}{4em}{\centering 3} & MonoSDF~\cite{Yu2022MonoSDF}  & N/A & 0.1020 & 0.3963 & 0.0830 & 0.4581 \\
    & Ours & 36 & \textbf{0.0115} & \textbf{0.8587} & \textbf{0.0124} & \textbf{0.8628} \\
    \bottomrule
  \end{NiceTabular}
  }
\vspace{-8pt}
\caption{Quantitative results on reconstructing 3D Shapes using the generated images.}
\vspace{-8pt}
  \label{tab:shape}
\end{table}

\paragraph{Multi-view Reconstruction} Given our approach can also synthesis novel views under the multi-view setting, we also show results where we use 3 conditional input views to synthesize 36 views for shape reconstruction. Compared with existing multi-view reconstruction frameworks such as MonoSDF~\cite{Yu2022MonoSDF}, our approach capture more details and generates smoother surfaces (note that~\cite{Yu2022MonoSDF} also relies on additional depth and normal estimations for its reconstruction whereas ours does not).

%% file: sec/6_ablation.tex
\begin{figure}
\centering
    \includegraphics[trim={0 1cm 0 0}, clip, width=0.99\linewidth]{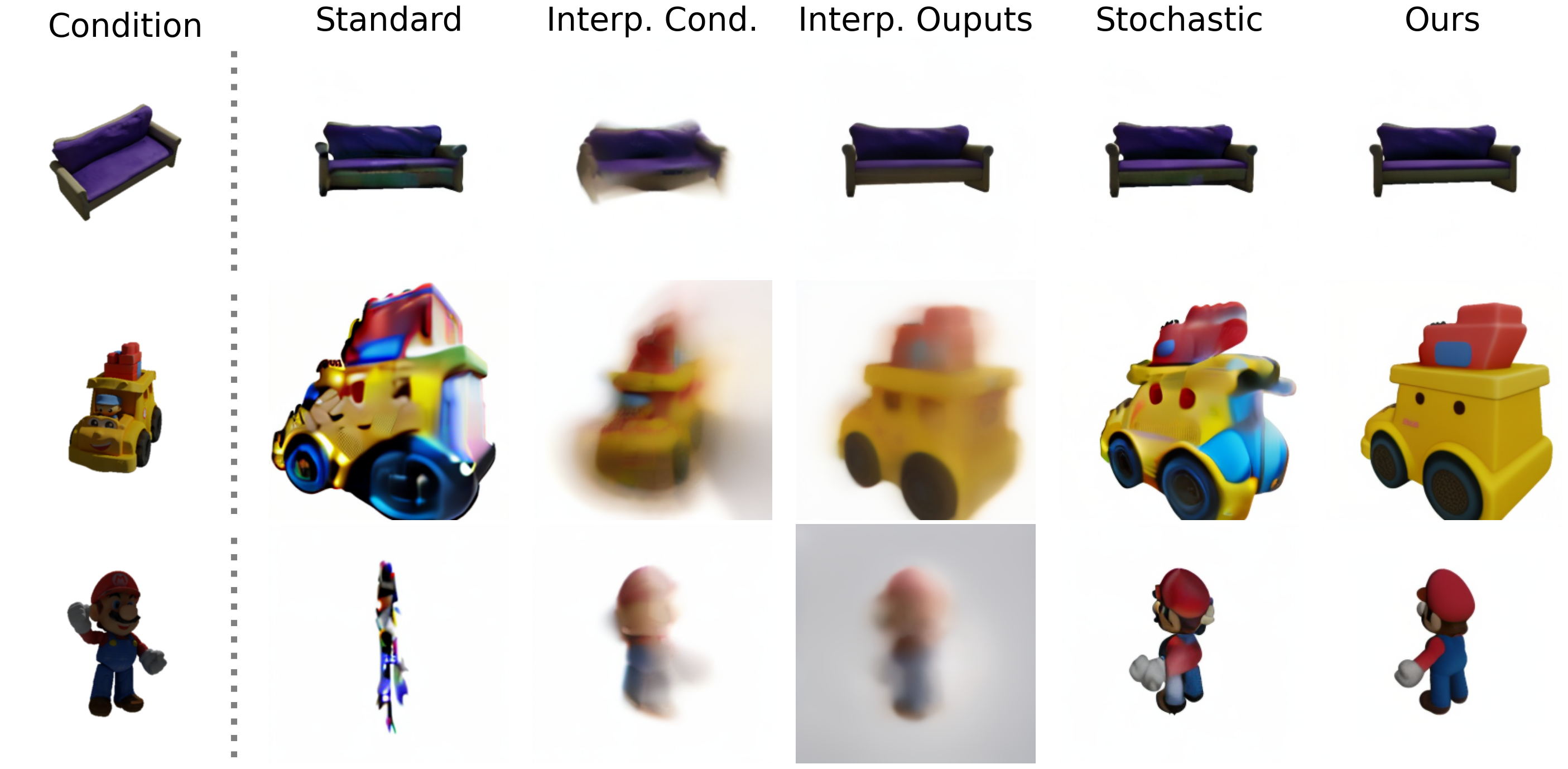}
\vspace{-8pt}
  \caption{Qualitative comparison for different auto-regressive generations.}
  \label{fig:ablation}
  \vspace{-10pt}
\end{figure}

\begin{table}
  \centering
  \resizebox{0.9\linewidth}{!}{
  \begin{tabular}{@{}lcccccc@{}}
    \toprule
    Dataset & Method & SSIM$\uparrow$ & PSNR$\uparrow$ & LPIPS$\downarrow$ \\
    \midrule
    \multirow{5}{3em}{ABO} & Zero123 (no interp.) & 0.8796 & 21.33 & 0.0961 \\
    & Standard auto-regression & 0.8010 & 16.77 & 0.1854\\
    &  Interpolated conditions  & 0.7243 & 13.26 & 0.3770 \\
    &  Interpolated outputs & \textbf{0.8925}* & \textbf{21.95}* & 0.1246 \\
    &  Stochastic conditioning~\cite{watson2022novel} & 0.8699 & 20.64 & 0.1106  \\
    &  Interpolated denoising  &  0.8848 & 21.43 & \textbf{0.0923} \\
    \cline{1-5}
    \multirow{5}{3em}{GSO} & Zero123 (no interp.) & 0.8710 & 20.33 & 0.1029  \\
    & Standard auto-regression & 0.8094 & 16.30 & 0.1801\\
    &  Interpolated conditions  & 0.7661 & 14.38 & 0.3427\\
    &  Interpolated outputs & 0.8799 & 20.44 & 0.1659\\
    &  Stochastic conditioning~\cite{watson2022novel} & 0.8658 & 19.97 & 0.1119 \\
    &  Interpolated denoising  & \textbf{0.8820} & \textbf{20.73} & \textbf{0.0958} \\
    \bottomrule
  \end{tabular}
  }
  \vspace{-8pt}
  \caption{Quantitative comparison between our method to different auto-regressive variants. Note that the \textit{Interpolated outputs} variant achieves the highest SSIM and PSNR values on the ABO dataset, but generates blurred images.}
  \label{tab:ablation}
  \vspace{-12pt}
\end{table}

\subsection{Ablations}
Here, we study the effectiveness of our proposed interpolated denoising process by exploring various auto-regressive generation variants (in~\cref{tab:ablation} and~\cref{fig:ablation}). Specifically, we investigate four variants of denoising: standard auto-regression, interpolated conditions, interpolated outputs, and stochastic conditioning which was proposed in~\cite{watson2022novel}. 


\paragraph{Standard Auto-regression.} One initial approach to auto-regression involves using the last generated view as the subsequent conditioning. However, this method exhibits bad generation quality, due to the accumulation of errors during the sequential generation process. As each subsequent view relies on the accuracy of the previous one, any inaccuracies or imperfections in earlier stages can compound, leading to a degradation in overall image quality. This limitation highlights the need for more sophisticated auto-regressive strategies to address the issue of error propagation and enhance the quality of generated views.

\paragraph{Interpolated Conditions and Interpolated Outputs.} Interpolated Conditions and Interpolated Outputs are two straightforward approaches to introduce auto-regressive generation into an existing diffusion model. The former method involves interpolating feature embeddings from condition images and poses, while the latter interpolates the final image feature maps produced by the model. Despite SSIM and PSNR metrics showing favorable results for Interpolated Outputs over others in~\cref{tab:ablation}, as illustrated in visual comparisons in~\cref{fig:ablation} shows that it leads to blurring of the output views and this is corroborated by larger LPIPS distance. 


\paragraph{Stochastic Conditioning.} We also explore the application of the Stochastic Conditioning Sampler proposed by~\cite{watson2022novel} to the Zero-1-to-3 model. We observe that Stochastic Conditioning fails to deliver the desired auto-regressive generation results; this may be attributed to the specific category on which the diffusion model used by~\citet{watson2022novel} was trained, allowing it to handle plausible condition images more effectively. By contrast, Zero-1-to-3~\cite{zero123} was trained on a cross-category dataset and designed for zero-shot reconstruction. Additionally, our evaluation data contains out-of-the-distribution data (\ie, ABO~\cite{abo} and GSO~\cite{downs2022google}).



%% file: sec/7_conclusion.tex
\section{Conclusion}


In this work, we have developed \modelname{}, a novel algorithm that addresses the challenge of multi-view consistency in novel-view synthesis with diffusion models. Our approach circumvents the need for fine-tuning or additional modules by integrating an auto-regressive mechanism that incrementally refines view synthesis, utilizing the entire history of previously generated views. Our proposed diffusion interpolation technique extends the denoising process in pre-trained diffusion models from a single-view setting to a multi-view setting without training requirements. Empirical evidence underscores \modelname{}'s capability to produce consistently high-quality views, and we achieve significant steps forward in novel view synthesis and 3D reconstruction applications.

%% file: sec/X_suppl.tex
\clearpage
\setcounter{page}{1}
\maketitlesupplementary


%

\begin{table*}
  \centering
  \begin{tabular}{@{}lcccc|ccc@{}}
    \toprule
    \multirow{2}{2em}{Dataset} & \multirow{2}{2em}{Method} & \multicolumn{3}{c|}{Image Quality} & \multicolumn{3}{c}{Multi-view Consistency} \\
     & & SSIM$\uparrow$ & PSNR$\uparrow$ & LPIPS$\downarrow$ & SIFT$\uparrow$ & LPIPS$\downarrow$ & CLIP$\uparrow$\\
    \midrule
    \multirow{7}{3em}{ABO} & Zero123 & 0.8796 & 21.33 & 0.0961 & 16.69 & 0.1234 & 0.9725\\
     & $\tau_c=0.33$ + $\tau_g=0.1$ & 0.8633 & 19.78 & 0.1168 & 18.32 & 0.0965 & 0.9804\\
     & $\tau_c=0.33$ + $\tau_g=0.5$ & 0.8788 & 20.86 & 0.0984 & 17.95 & \textbf{0.0945} & \textbf{0.9815}\\
     & $\tau_c=0.33$ + $\tau_g=1.0$ & 0.8804 & 21.06 & 0.0961 & 17.94 & 0.0948 & 0.9812\\
     & $\tau_c=0.50$ + $\tau_g=0.1$ & 0.8753 & 20.56 & 0.1045 & \textbf{18.46} & 0.0968 & 0.9813\\
     & $\tau_c=0.50$ + $\tau_g=0.5$ & \textbf{0.8848} & 21.35 & 0.0933 & 18.12 &  0.0964 & 0.9813\\
     & $\tau_c=0.50$ + $\tau_g=1.0$ & 0.8848 & \textbf{21.43} & \textbf{0.0923} & 18.01 & 0.0966 & 0.9812\\
    \bottomrule
    \multirow{7}{3em}{GSO} & Zero123 & 0.8710 & 20.33 & 0.1029 & 15.15 & 0.1054 & 0.9592\\
    & $\tau_c=0.33$ + $\tau_g=0.1$ & 0.8632 & 19.15 & 0.1193 & \textbf{19.43} & 0.0675 & 0.9760\\
     & $\tau_c=0.33$ + $\tau_g=0.5$ & 0.8770 & 20.18 & 0.1020 & 18.64 & \textbf{0.0664} & \textbf{0.9779}\\
     & $\tau_c=0.33$ + $\tau_g=1.0$ & 0.8789 & 20.38 & 0.0994 & 18.54 & 0.0671 & 0.9778\\
     & $\tau_c=0.50$ + $\tau_g=0.1$ & 0.8725 & 19.89 & 0.1081 & 19.13 & 0.0675 & 0.9764\\
     & $\tau_c=0.50$ + $\tau_g=0.5$ & 0.8812 & 20.62 & 0.0969 & 18.30 & 0.0689 & 0.9773\\
     & $\tau_c=0.50$ + $\tau_g=1.0$ & \textbf{0.8820} & \textbf{20.73} & \textbf{0.0958} & 17.95 & 0.0676 & 0.9773 \\
    \bottomrule
  \end{tabular}
  \caption{Experiments about condition image weights.}
  \label{tab:parameter}
\end{table*}



\section{Implementation}

We implement our auto-regressive techniques on the pre-trained Zero-1-to-3~\cite{zero123}. To facilitate single-view generation and spin video generation, we set a maximum offset per step, denoted as $\delta=10^\circ$ for most cases except 16 view spin video generation. For a fair comparison with SyncDreamer, we modify our setup to match their conditions, where $\delta=22.5^\circ$ to generate 16 view images, aligning with SyncDreamer's configuration. We have conducted an investigation into various values for the temperature parameters, $\tau_c$ and $\tau_g$, in Eqs.(\textcolor{red}{12}) and (\textcolor{red}{13}). Our experiments reveal that setting $\tau_c$ to 0.5 and $\tau_g$ to 1.0 leads to superior results, as evidenced by the data presented in~\cref{tab:parameter}. The \emph{Interpolated Denoising} process is illustrated in~\cref{alg:interp}. For reconstruction, we optimize the NeuS~\cite{wang2021neus} using the generated multi-view images with their corresponding masks from Zero-1-to-3~\cite{zero123}, SyncDreamer~\cite{syncdreamer} and ours. For One-2-3-45~\cite{one2345}, we directly follow the their pipeline, which requires elevation estimation. To apply \modelname{} on in-the-wild images, we apply an off-the-shelf background removal tool CarveKit to remove the background and adjust the object ratio on the image.

\begin{algorithm}
\caption{Interpolated denoising with classifier-free guidance}\label{alg:interp}
\begin{algorithmic}
\State Input: conidition $y$, unconditional scale $u$, $\alpha_t$, $\sigma_t$, $\tau_c$, $\tau_g$
\State Determine generated trajectory ${x_0^1,x_0^2,...,x_0^N}$
\State Add $x_0^1\gets y$ to condition set
\State $\{w_1,w_2...,w_N\} \gets Eq.\textcolor{red}{13}$
\For{$n$ from $2$ to $N$}
    \State $x_T \gets $ Sample from $\mathcal{N}(\mathbf{0}, \mathbf{I})$
    \For{$t$ from $T$ to $1$}
        \State $y^i \gets$ Sample $x_0^i$ from condition set
        \State $\epsilon_t^i \gets \epsilon_t^i(x_t, \emptyset) + u\left(\epsilon_t^i(x_t, y^i) - \epsilon_t^i(x_t, \emptyset)\right)$
        \State $\epsilon' \gets$ Sample from $\mathcal{N}(\mathbf{0}, \mathbf{I})$
        \State $x_{t-1}^i \gets  \sqrt{\bar{\alpha}_{t-1}}\left(\frac{x_t-\sqrt{1-\bar{\alpha}_{t-1}}\epsilon_t}{\sqrt{\bar{\alpha}_t}}\right)+\sqrt{1-\bar{\alpha}_{t-1}-\sigma_t^2}\cdot\epsilon_t^i+\sigma_t\epsilon'$
        \State $x_{t-1}^n \gets \sum_{i=1}^n \omega_i x_{t-1}^i$
    \EndFor
    \State Add $x_0^n$ to condition set
\EndFor
\end{algorithmic}
\end{algorithm}

\section{Multi-view generation.}
We formulate the weights to single-view generation in Eq~\red{13}. In the general case, when given k views, the weights are expressed as follows,
\begin{equation}
\small
\label{eq:weights}
\omega_n=\left\{
\begin{aligned}
    &exp(-\frac{\Delta^{n}}{\tau_c})\text{Softmax}(\frac{e^{-\frac{\Delta^{n}}{\tau_c}}}{\sum_{n=1}^k e^{-\frac{\Delta^n}{\tau_c}}}) & n=1, \ldots, k\\
    &(1-\sum_{i=1}^{k}\omega_i)\text{Softmax}(\frac{e^{-\frac{\Delta^{n}}{\tau_g}}}{\sum_{n=k+1}^N e^{-\frac{\Delta^n}{\tau_g}}}), & n > k
\end{aligned}
\right.
\end{equation}
where we apply the term $1-\sum_{i=1}^{k}\omega_i$ on the generated image weights to ensure sum of all weights equals $1$ as a requirement for the objective $\sum_{n=1}^{N}{w_n}=1$.

\section{Image Rendering}
We organize the testing data by using the rendering scripts provided by both Zero-1-to-3 and SyncDreamer respectively. It's important to note that there are slight variations in the camera and lighting settings between the two approaches.

\noindent\textbf{Camera.} Zero-1-to-3 employs random sampling for the camera distance within a range of $[1.5, 2.2]$. The azimuth and elevation angles for both condition and target images are randomly selected. SyncDreamer maintains a fixed camera distance of 1.5 and samples azimuth angles from a discrete angle set $\{0^\circ, 22.5^\circ, 45^\circ, ..., 337.5^\circ\}$ for both condition and target images. The condition elevation is randomly sampled within the range of $[0^\circ, 30^\circ]$, while the target elevation is fixed at $30^\circ$.

\noindent\textbf{Lighting.} Zero-1-to-3 uses point light as its lighting model. SyncDreamer, on the other hand, employs a uniform environment light setup. This choice of lighting leads to differences in the rendering results. Specifically, renderings from Zero-1-to-3 exhibit shadows on the backside of the objects, whereas those from SyncDreamer do not. 

These discrepancies in rendering impact the evaluation of 3D reconstructions. As we take Zero-1-to-3 as our baseline, we adopt the consistent rendering settings with Zero-1-to-3 to organize test data for fair comparison.

\begin{figure}[t]
    \includegraphics[width=0.99\linewidth]{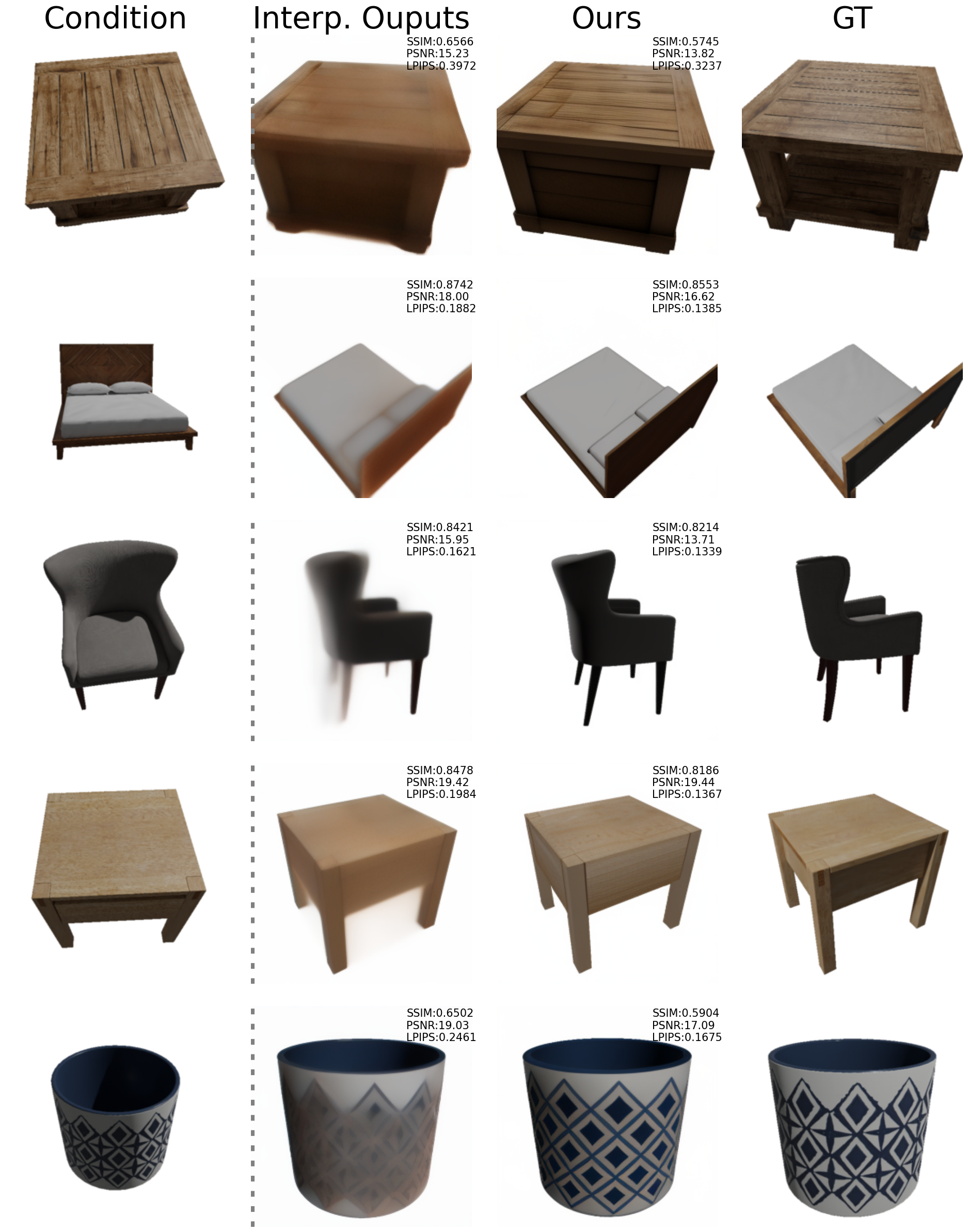}
    \caption{Visual comparison for SSIM and PSNR limitation in capturing blur.}
  \label{fig:ssim_psnr_fail}
\end{figure}
\section{SSIM and PSNR}

In the main manuscript, we mentioned the limitations of SSIM and PSNR in effectively capturing blur, as detailed in~\cref{tab:ablation}. We further underscore these limitations with illustrative examples, as depicted in~\cref{fig:ssim_psnr_fail}, where images with higher SSIM and PSNR scores exhibit pronounced blurriness. Our findings highlight the comparative shortcomings of output interpolation when compared with diffusion interpolation. Importantly, we stress that LPIPS provides a more precise assessment of image quality.

\begin{figure}[t]
    \includegraphics[width=0.99\linewidth]{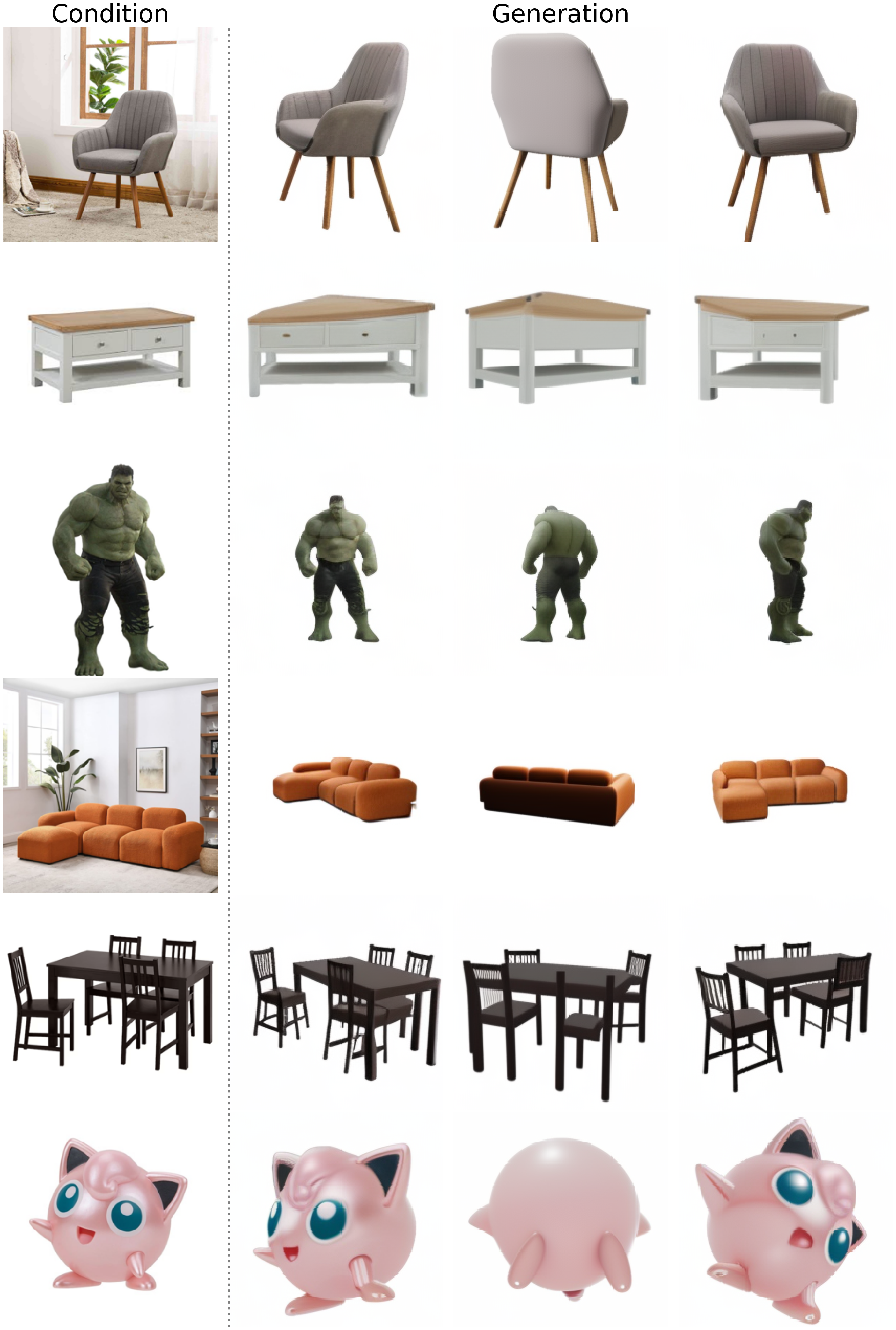}
    \caption{Visual examples for failure cases. The failure cases mainly includes failure under specific views (1st and 2nd rows), face (3rd row), detailed textures (4th row), complex scenes (5th row), and elevation angle ambiguity (6th row)}
  \label{fig:fail}
\end{figure}
\section{Limitation}
While our model, ~\modelname{}, demonstrates promising performance in significantly enhancing the multi-view consistency of the original Zero-1-to-3 framework, there are certain limitations that remain unaddressed by the current framework.

First, ~\modelname{} relies on using all generated images to guide the generation process. This requirement necessitates additional memory to store these images and imposes a sequential nature to the generation process. In contrast, the original Zero-1-to-3 can proceed spin video generation as a batch process and generate views in parallel, resulting in a more time-efficient approach. The sequential generation nature of ~\modelname{} can lead to additional time consumption. Considering to generate a single image, Zero-1-to-3 takes roughly 4s, while our methods takes $4s\sim45s$ (from 1 condition to 24 conditions) depending on the size of the condition set.

Second, \modelname{} heavily relies on the pre-trained Zero-1-to-3 model. While it is generally effective, there are still instances where it fails, particularly under certain specific views. Even with the integration of auto-regressive generation, it cannot entirely mitigate this limitation, as demonstrated in the~\cref{fig:fail} 1st and 2nd examples.

Third, although current pose-conditional diffsuion models~\cite{zero123,syncdreamer,tang2023mvdiffusion,tseng2023consistent,weng2023consistent123,yang2023consistnet} have been trained on large-scale 3D dataset, \ie., Objaverse~\cite{deitke2023objaverse,deitke2023objaversexl}, they are still struggling to deal with scenes that comprise intricate details (\eg, human faces, detailed textures) as shown by the 3rd and 4th examples in the~\cref{fig:fail}, complex scenes, as shown by 5th examples in~\cref{fig:fail}, and the models may struggle with elevation angle ambiguity, as demonstrated by the 6th example in~\cref{fig:fail}. In these cases, the model's performance may be limited in capturing all the fine-grained information and nuances.

\section{Application}

\begin{sidewaysfigure*}
    \centering
    \includegraphics[width=1.01\linewidth]{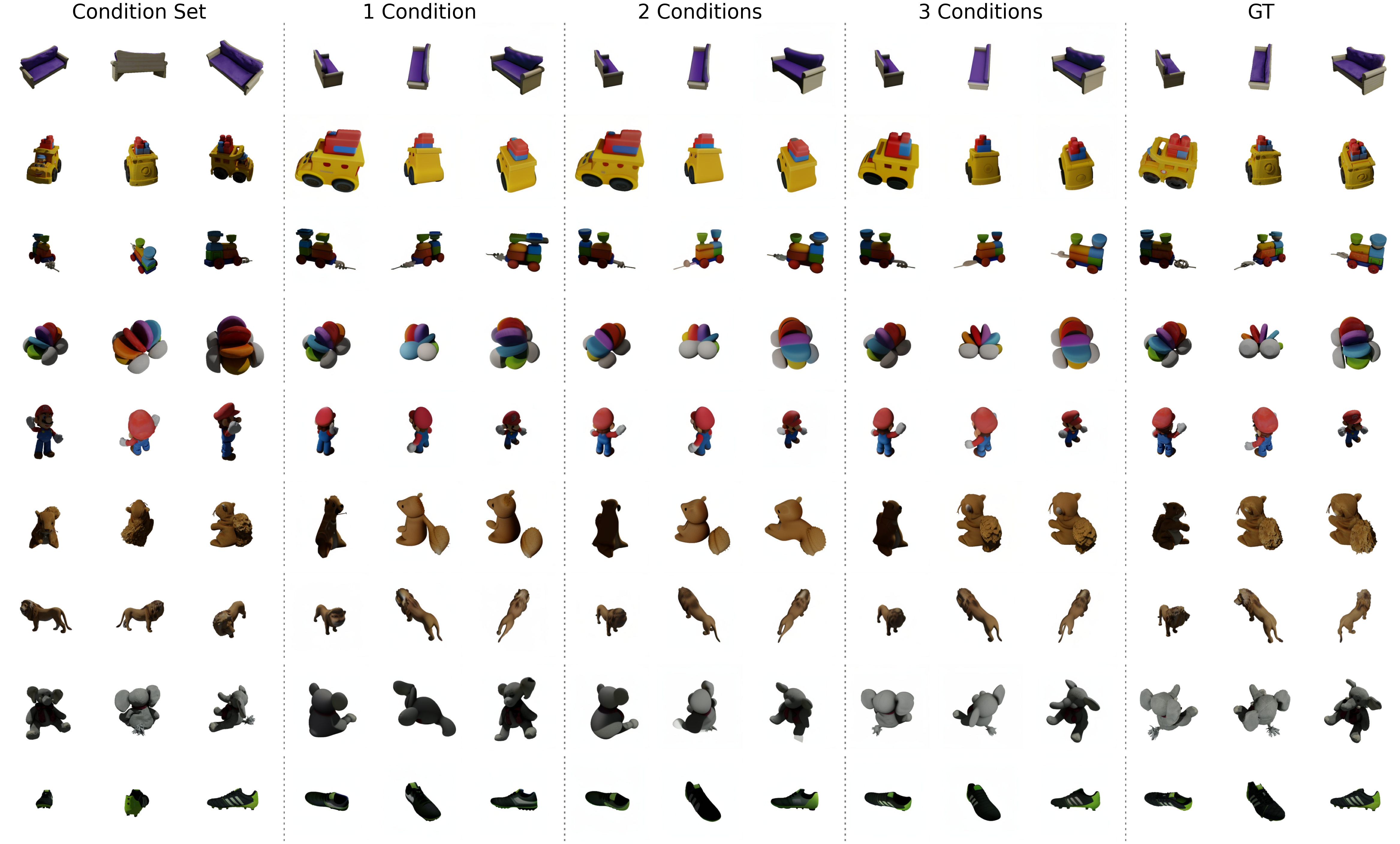}
    \caption{Qualitative comparison for Multi-view reconstruction.}
    \label{fig:multiview}
\end{sidewaysfigure*}

\textbf{Multi-view generation.} As mentioned in the main manuscript, thanks to the multi-view conditioned ability by the introduced interpolated denoising process, we could extend the single-view conditioned model into multi-view conditioned model easily, thus enabling support for multi-view reconstruction. The quantities results presented in~\cref{tab:multi_view} and we provide qualitative comparison in~\cref{fig:multiview} here to further demonstrate the advantages of our method, as it consistently yields improved reconstructions with an increasing number of views. This clear improvement demonstrate the effectiveness of our proposed techniques in handling multi-view condition images.

\noindent\textbf{Consistent BRDF decomposition.} In our experimental observations, we identified a particular challenge encountered by the pre-trained decoder, which often struggles to effectively distinguish between shadows and surface textures in images. To overcome this limitation, we introduced a dedicated decomposition decoder, specifically designed to meticulously separate these visual elements. When this decomposition decoder is integrated with our interpolated denoising approach, it not only upholds multi-view consistency but also exhibits the potential to excel in novel-view decomposition and rendering tasks.

\begin{figure*}[t]
\centering
    \includegraphics[width=0.85\linewidth]{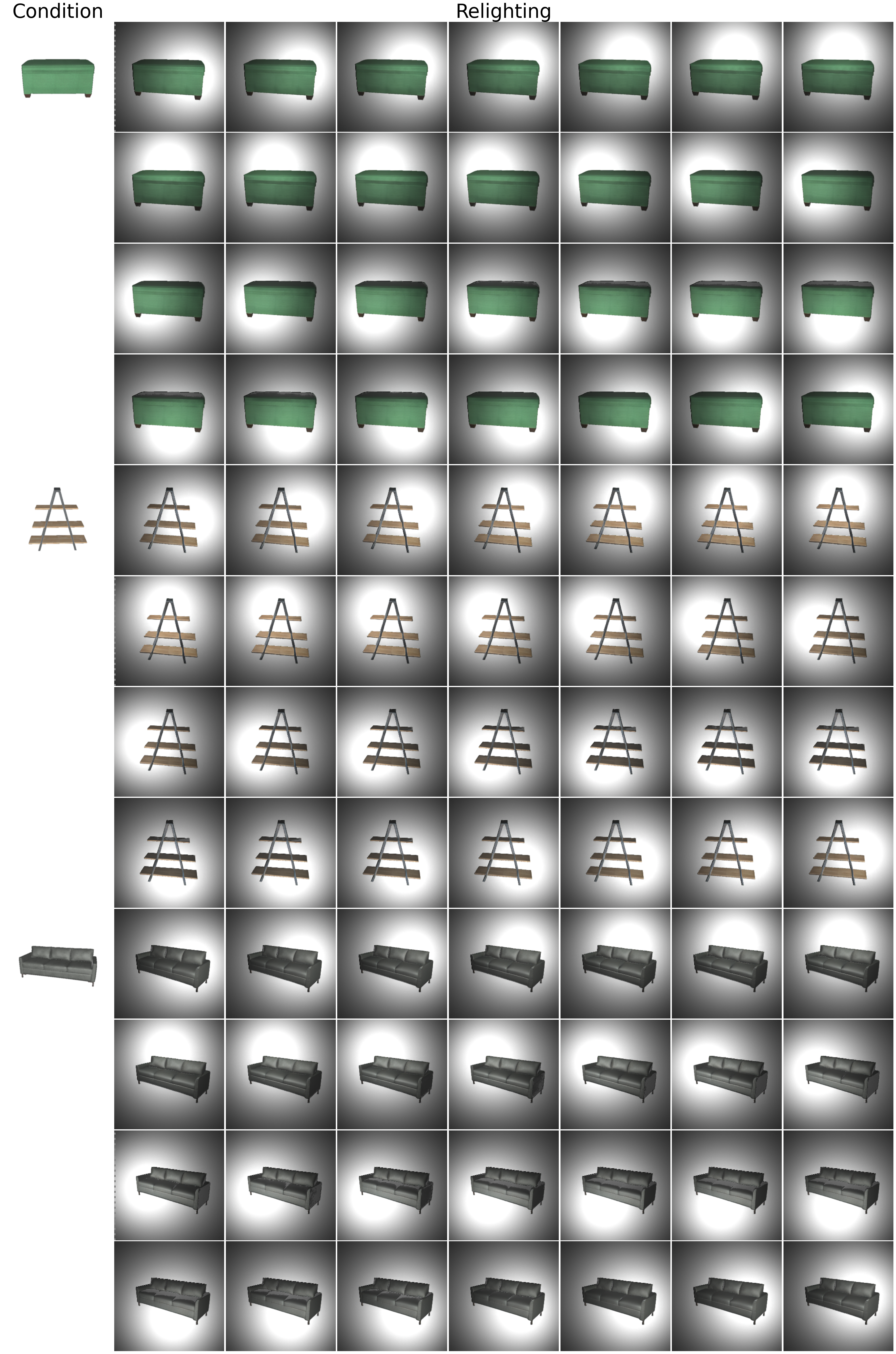}
    \caption{Qualitative comparison for relighting.}
  \label{fig:relighting}
\end{figure*}

This novel combination of techniques offers promising possibilities. By leveraging decomposed BRDF (Bidirectional Reflectance Distribution Function) maps, we gain greater control over the lighting and shape geometry of the scenes. The availability of normal maps enhances our ability to manipulate the lighting conditions, promising more flexibility in rendering as shown in~\cref{fig:relighting}. With this level of control, we can explore various exciting applications, such as dynamic relighting, creative scene composition, and the generation of captivating visual effects. This opens up new avenues for artistic and practical image and video manipulation, granting artists and professionals the tools to craft engaging and visually stunning content.

\begin{figure*}[t]
    \includegraphics[width=0.99\linewidth]{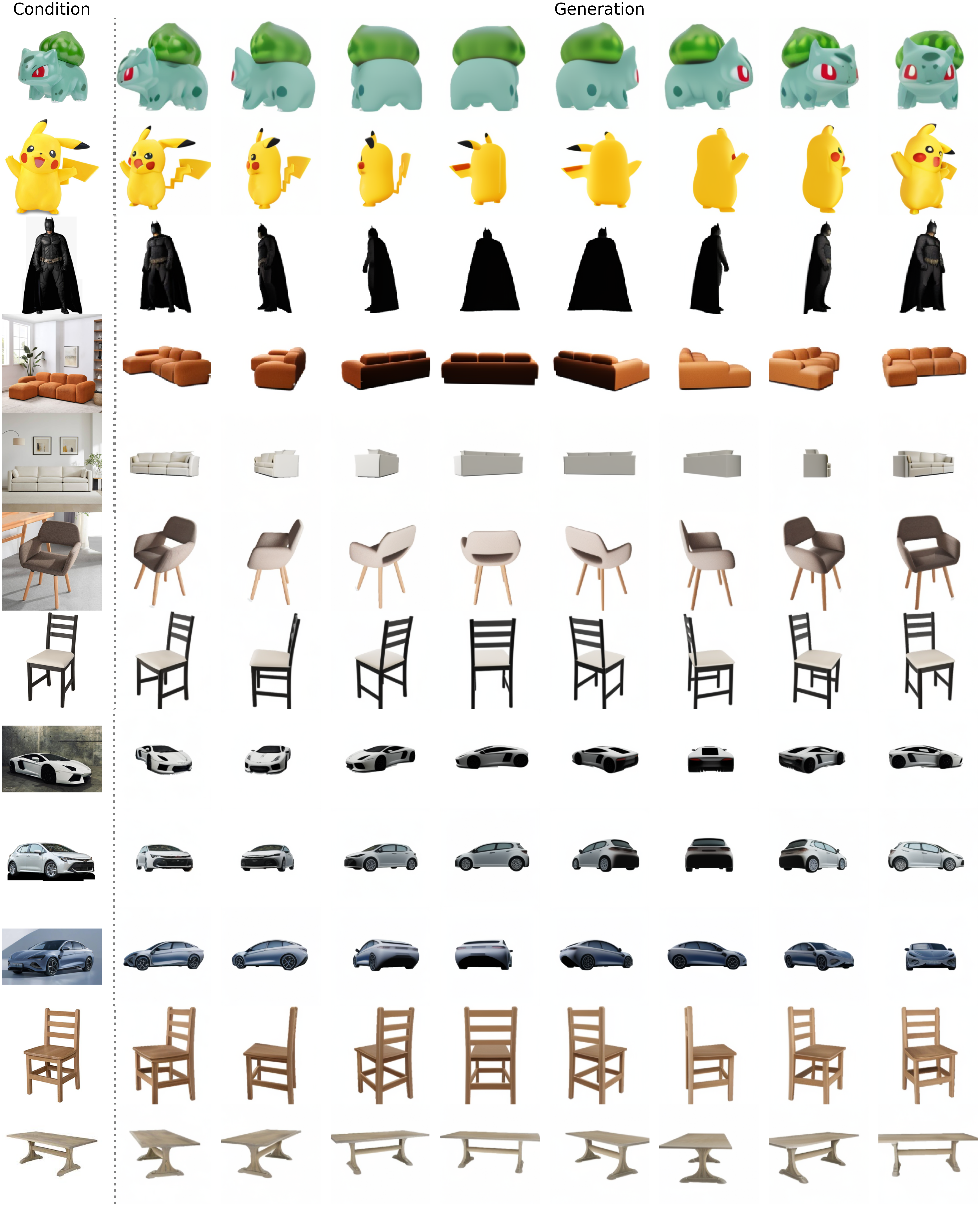}
  \caption{Visualization on real images. Images were downloaded online, where foreground objects were segmented and the image was resized to be aligned with pre-training images.}
  \label{fig:real}
\end{figure*}

\begin{figure*}[t]
    \includegraphics[width=0.99\linewidth]{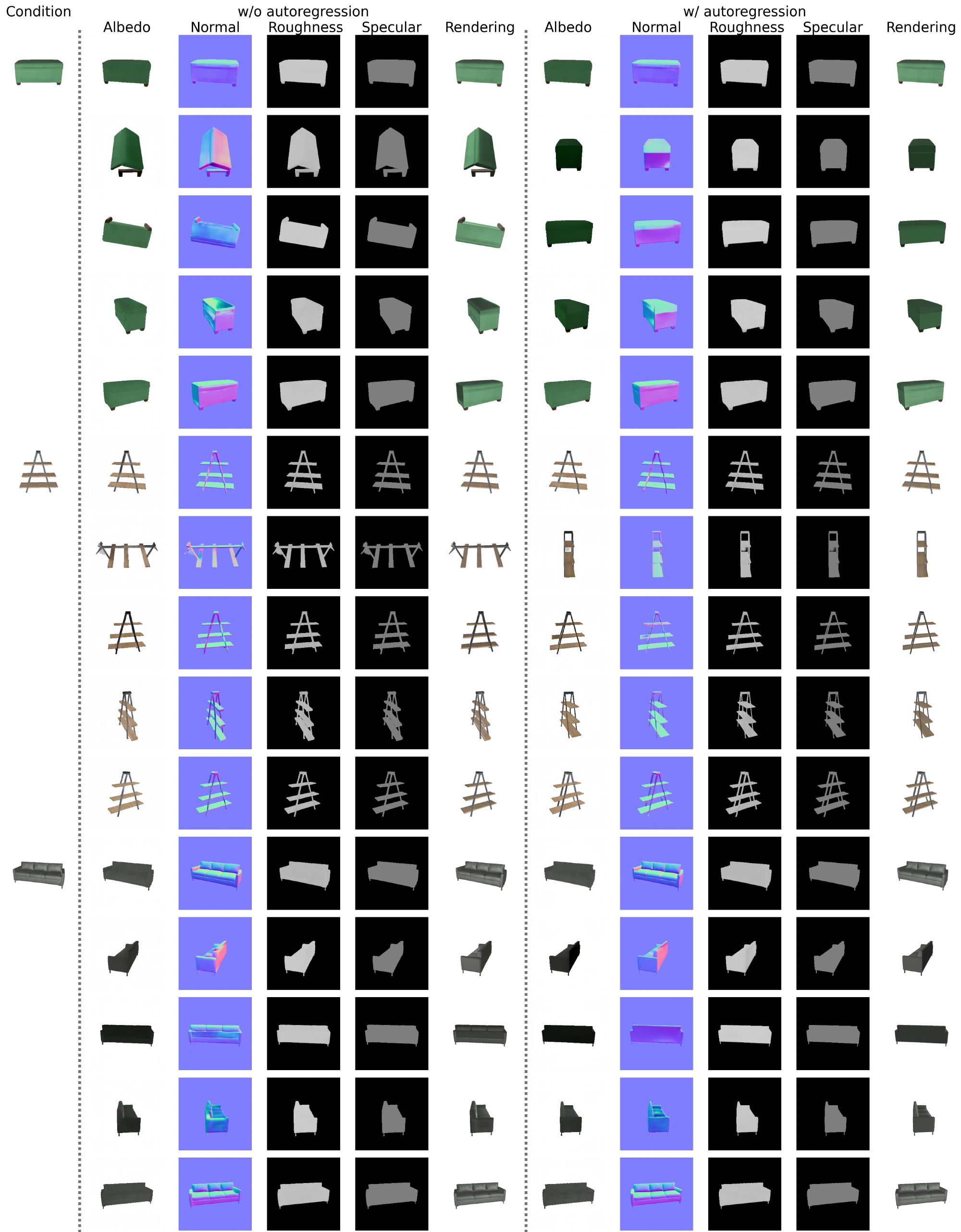}
    \caption{Qualitative comparison for BRDF decomposition w/o vs w/ autoregression.}
  \label{fig:decomposition}
\end{figure*}

